\documentclass{article}

\PassOptionsToPackage{numbers, compress}{natbib}

\usepackage[final]{neurips_2020} %

\usepackage[utf8]{inputenc} %
\usepackage[T1]{fontenc}    %
\usepackage{hyperref}       %
\usepackage{url}            %
\usepackage{booktabs}       %
\usepackage{amsfonts}       %
\usepackage{nicefrac}       %
\usepackage{microtype}      %
\usepackage{tablefootnote}

\usepackage{etoolbox} %
\usepackage{amsmath}
\usepackage{amssymb}
\usepackage{graphicx}
\usepackage{subcaption}
\usepackage{listings}
\usepackage{float}
\usepackage{enumitem}
\usepackage{xspace}
\usepackage{multirow}
\usepackage{array}
\usepackage{tabularx}
\usepackage{booktabs}
\usepackage{xcolor,colortbl}
\usepackage{cleveref}
\usepackage[english]{babel}
\usepackage{pifont} 
\usepackage{soul}
\usepackage{slashbox}
\usepackage{xfrac}
\usepackage{wrapfig}
\usepackage{mdwlist}
\usepackage{arydshln}

\usepackage{adjustbox}

\usepackage[ruled,vlined]{algorithm2e}

\newtheorem{definition}{Definition}

\newtheorem{theorem}{Theorem}

\newtheorem{proof}{Proof}

\crefformat{section}{\S#2#1#3} %
\crefformat{subsection}{\S#2#1#3}
\crefformat{subsubsection}{\S#2#1#3}
\newcolumntype{L}[1]{>{\raggedright\let\newline\\\arraybackslash\hspace{0pt}}m{#1}}
\newcolumntype{C}[1]{>{\centering\let\newline\\\arraybackslash\hspace{0pt}}m{#1}}
\newcolumntype{R}[1]{>{\raggedleft\let\newline\\\arraybackslash\hspace{0pt}}m{#1}}

\SetCommentSty{mycommfont}

\definecolor{purple2}{RGB}{153,0,153} %
\definecolor{green2}{RGB}{0,153,0} %
\newcommand{\method}{\textsc{H$_{2}$GCN}\xspace}
\newcommand{\stepone}{\textbf{{(S1)}}\xspace}
\newcommand{\steptwo}{\textbf{{(S2)}}\xspace}
\newcommand{\stepthree}{\textbf{{(S3)}}\xspace}

\newcommand{\mechanismOne}{Ego- and Neighbor-embedding Separation}
\newcommand{\mechanismTwo}{Higher-order Neighborhoods}
\newcommand{\mechanismThree}{Combination of Intermediate Representations}

\newcommand{\mechanismOneSmall}{ego- and neighbor-embedding separation}
\newcommand{\mechanismTwoSmall}{higher-order neighborhoods}
\newcommand{\mechanismThreeSmall}{combination of intermediate representations}

\newcommand{\mymathhl}[1]{\colorbox{gray!15}{$\displaystyle #1$}}
\newcommand{\mytexthl}[1]{\colorbox{gray!15}{#1}}

\definecolor{darkgreen}{RGB}{0,153,0}

\newcommand{\cmark}{{\color{darkgreen}\ding{51}}}
\newcommand{\xmark}{\ding{55}}

\lstdefinestyle{common}{
	basicstyle = \ttfamily,
	keywordstyle=\color{blue},       %
	keywordstyle={[2]\color{cyan}}, %
	stringstyle=\color{purple2},
	commentstyle=\color{green2},
	upquote=true,                  %
	breaklines=true, frame=trBL
}

\newlist{steps}{enumerate}{1}
\setlist[steps, 1]{label = Step \arabic*:}

\newcommand{\V}[1]{\mathbf{#1}}
\newcommand{\R}{\mathbb{R}}

\newcommand{\graph}{\mathcal{G}}
\newcommand{\vertexSet}{\mathcal{V}}
\newcommand{\edgeSet}{\mathcal{E}}

\newcommand{\matA}{\mathbf{A}}
\newcommand{\matB}{\mathbf{B}}
\newcommand{\matAh}{\mathbf{\hat{A}}}
\newcommand{\matD}{\mathbf{D}}
\newcommand{\matDh}{\mathbf{\hat{D}}}

\newcommand{\matX}{\mathbf{X}}

\newcommand{\matW}{\mathbf{W}}
\newcommand{\matH}{\mathbf{H}}
\newcommand{\matI}{\mathbf{I}}
\newcommand{\matL}{\mathbf{L}}
\newcommand{\matR}{\mathbf{R}}

\newcommand{\vecy}{\mathbf{y}}

\newcommand{\setY}{\mathcal{Y}}
\newcommand{\setT}{\mathcal{T}_{\vertexSet}}
\newcommand{\T}{\mathsf{T}}

\newcommand{\neighNoSelfLoop}{\bar{N}}

\newcolumntype{H}{>{\setbox0=\hbox\bgroup}c<{\egroup}@{}}

\makeatletter
\newcommand\footnoteref[1]{\protected@xdef\@thefnmark{\ref{#1}}\@footnotemark}
\makeatother
\ifdefmacro{\inpreprint}
{\title{Generalizing Graph Neural Networks Beyond Homophily}} %
{\title{Beyond Homophily in Graph Neural Networks: Current Limitations and Effective Designs}} %

\author{%
  Jiong Zhu \\
  University of Michigan\\
  \texttt{jiongzhu@umich.edu} \\
  \And
  Yujun Yan \\
  University of Michigan \\
  \texttt{yujunyan@umich.edu} \\
  \And
  Lingxiao Zhao \\
  Carnegie Mellon University \\
  \texttt{lingxia1@andrew.cmu.edu} \\
  \And
  Mark Heimann \\
  University of Michigan\\
  \texttt{mheimann@umich.edu} \\
  \And
  Leman Akoglu \\
  Carnegie Mellon University \\
  \texttt{lakoglu@andrew.cmu.edu} \\
  \And
  Danai Koutra \\
  University of Michigan\\
  \texttt{dkoutra@umich.edu}
}

\begin{document}

\maketitle

\begin{abstract}

We investigate the representation power of graph neural networks in the semi-supervised node classification task under \textit{heterophily} or \textit{low homophily}, i.e., in networks where 
connected nodes may have \textit{different} class labels and \textit{dissimilar} features.
Many popular %
GNNs fail to generalize to this setting, and are even outperformed by models that ignore the graph structure (e.g., multilayer perceptrons). %
Motivated by this limitation, we identify a set of key designs---\mechanismOneSmall, %
\mechanismTwoSmall, and \mechanismThreeSmall---that boost  
learning from the graph structure under heterophily.
{We combine them into a graph  neural network, \method, which we use as the base method %
to empirically evaluate the effectiveness of the identified designs.}
{Going beyond the traditional benchmarks with strong homophily, 
our empirical analysis 
shows that the identified designs
increase the accuracy of GNNs
by {up to 40\% and 27\% 
over models without them 
on synthetic and real networks with heterophily, respectively,} %
{and yield} %
competitive performance under homophily.}

\end{abstract}

\section{Introduction}

We focus on the effectiveness of graph neural networks (GNNs)~\cite{dlgsurvey_tkde20} 
in tackling the semi-supervised node classification task  in challenging %
settings:
the goal of the task is to infer the unknown labels of the nodes by using the network structure~\cite{zhu2005},
given partially labeled networks with node features (or attributes).
Unlike most prior work that considers networks with strong homophily, we study the representation power of GNNs in settings with \textit{different levels of homophily} or \textit{class label smoothness}.

Homophily is a key principle of many real-world networks, 
whereby linked nodes often belong to the same class or have similar features (``birds of a feather flock together'')~\cite{mcpherson2001birds}. 
For example, friends are likely to have similar political beliefs or age, and papers tend to cite papers from the same %
research area~\cite{newman2018networks}. 
GNNs \textit{model the homophily principle} by propagating features and aggregating them within various graph neighborhoods via different mechanisms (e.g., averaging, LSTM)~\cite{kipf2016semi,hamilton2017inductive,velickovic2018graph}.  
However, in the real world, there are also settings where ``opposites attract'', leading to networks with \textit{heterophily}: linked nodes are likely from different classes or have dissimilar features. For instance, the majority of people tend to connect with people of the opposite gender in dating networks,  different amino acid types are more likely to connect in protein structures, fraudsters are more likely to connect to accomplices than to other fraudsters in online purchasing networks~\cite{netprobe07}. %

Since {many existing} GNNs assume strong homophily, {they} fail to generalize
to networks with heterophily (or low/medium level of homophily).
In such cases, we find that even models that ignore the graph structure altogether, such as multilayer perceptrons or MLPs, can outperform a number of existing GNNs.
Motivated by this limitation,  we make the following contributions:
\setlist{leftmargin=*}
\begin{itemize*}
\item \textbf{Current Limitations}: We reveal the limitation of GNNs to learn over  networks with heterophily, which is ignored in the literature due to  evaluation on few benchmarks with similar properties. \S~\ref{sec:design}
\item \textbf{Key Designs for Heterophily \& New Model:} %
We identify a set of key designs that can boost learning from the graph structure in heterophily without trading off accuracy in homophily: (D1)~\mechanismOneSmall, (D2)~\mechanismTwoSmall, and (D3)~\mechanismThreeSmall. We justify the designs theoretically, and combine them into a model, \method, 
that effectively adapts to both heterophily and homophily.
{We compare it to prior GNN models, and make our code and data available at \url{https://github.com/GemsLab/H2GCN}.} 
\S~\ref{sec:design}-\ref{sec:related} %
\item \textbf{Extensive Empirical Evaluation}: 
We empirically analyze our model and competitive existing GNN models on both synthetic and real networks covering the full spectrum of low-to-high homophily
(besides the typically-used benchmarks with strong homophily only). %
{In synthetic networks, our detailed ablation study of \method (which is free of confounding designs) shows that the identified designs result in up to 40\% performance gain in heterophily. In real networks, we observe that GNN models utilizing even a subset of our identified designs outperform popular models without them by up to 27\% in heterophily, while being competitive in homophily.}
\S~\ref{sec:exp} 

\end{itemize*}

\vspace{-0.15cm}
\section{Notation and Preliminaries} %
\label{sec:preliminaries}
\vspace{-0.1cm}

\begin{wrapfigure}{r}{0.27\textwidth}
\vspace{-1.25cm}
\centering
\includegraphics[width=0.25\textwidth]{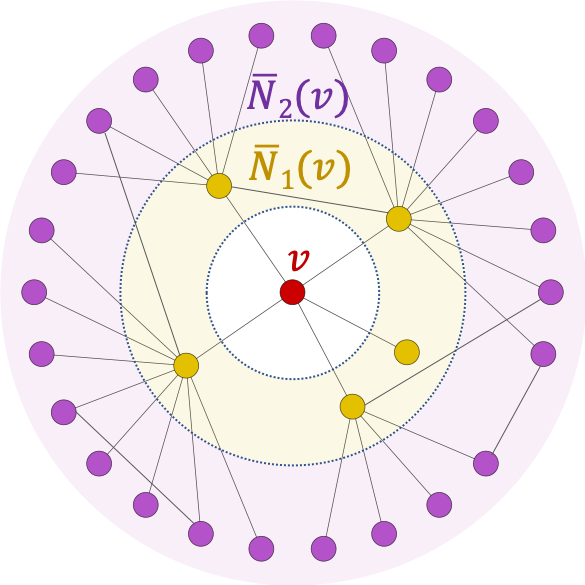}
\caption{Neighborhoods.}
\label{fig:neighborhoods}
\vspace{-0.3cm}
\end{wrapfigure}
We summarize our notation in Table~\ref{tab:dfn} (App.~\ref{app:dfn}).
Let $\graph=(\vertexSet,\edgeSet)$ be an undirected, unweighted graph with nodeset $\vertexSet$ and edgeset $\edgeSet$. 
We denote a general neighborhood centered around $v$ as $N(v)$ ($\graph$ may have self-loops), the corresponding neighborhood that does \textit{not} include the ego (node $v$) as $\neighNoSelfLoop(v)$, and the general neighbors of node $v$ at exactly $i$ hops/steps away (minimum distance) as $N_i(v)$. For example, $N_1(v) = \{u: (u,v) \in \edgeSet \}$ are the immediate neighbors of $v$. Other examples are shown in Fig.~\ref{fig:neighborhoods}.
We represent the graph by 
its adjacency matrix $\matA \in \{0,1\}^{n\times n}$ 
and its node feature matrix $\matX \in \mathbb{R}^{n \times F}$, where 
the vector $\V{x}_v$ corresponds to the \textit{ego-feature} of node $v$, 
and $\{\V{x}_u: u \in \neighNoSelfLoop(v)\}$ to its \textit{neighbor-features}.

We further assume a class label vector $\vecy$, which for each node $v$ contains a unique class label $y_v$.
The goal of semi-supervised node classification is to learn a mapping $\ell: \vertexSet \rightarrow \setY$, where $\setY$ is the set of labels, given a set of labeled nodes $\setT = \{(v_1,y_1), (v_2, y_2), ...\}$ as training data.

\textbf{Graph neural networks $\;$} 
\label{sec:2-gnn}
From a probabilistic perspective, most GNN models assume the following local Markov property on node features: 
for each node $v \in \vertexSet$, there exists a neighborhood $N(v)$ such that
$y_v$ only depends on the ego-feature $\V{x}_v$ and neighbor-features $\{\V{x}_u: u \in N(v)\}$. 
Most models derive the class label $y_v$ via the following representation learning approach: %
{\small 
\begin{equation}
    \V{r}^{(k)}_v = f\left(\V{r}^{(k-1)}_v, \{\V{r}^{(k-1)}_u: u \in N(v)\}\right), %
    \; \V{r}^{(0)}_v = \V{x}_v, \; \text{and} \;  y_v = \arg \max \{\mathrm{softmax}(\V{r}^{(K)}_v)\matW\},
\end{equation}
}
where the embedding function $f$ is applied repeatedly in $K$ total rounds, 
node $v$'s representation (or hidden state vector) at round $k$, {\small $\V{r}^{(k)}_v$}, is learned from its ego- and neighbor-representations in the previous round, and a softmax classifier with {learnable weight matrix} $\matW$ is applied to the final representation of $v$.
Most existing models differ in their definitions of neighborhoods $N(v)$ and embedding function $f$. 
A typical definition of neighborhood is $N_1(v)$---i.e., the 1-hop neighbors of $v$. 
As for $f$, in graph convolutional networks (GCN)~\cite{kipf2016semi} %
each node repeatedly averages its own features and those of its neighbors to update its own feature representation.  Using an attention mechanism, GAT~\cite{velickovic2018graph} models the influence of different neighbors more precisely as a weighted average of the ego- and neighbor-features. GraphSAGE~\cite{hamilton2017inductive} generalizes the aggregation beyond averaging, and models the ego-features distinctly from the neighbor-features in its subsampled neighborhood.

\textbf{Homophily and heterophily $\;$} 
\label{sec:homophily}
In this work, we focus on heterophily in class labels. %
We first define the edge homophily ratio $h$ as a measure of the graph homophily level, and use it to define graphs with strong homophily/heterophily: 
\begin{definition}
\vspace{-0.15cm}
The edge homophily ratio $h = \tfrac{|\{(u,v): (u,v) \in \edgeSet \wedge y_u = y_v\}|}{|\edgeSet|}$ is the fraction of edges in a graph which connect nodes that have the same class label (i.e., intra-class edges). 
\label{dfn:homophily-ratio}
\end{definition}
\begin{definition} %
\label{dfn:heterophily-nets}
\vspace{-0.15cm}
Graphs with strong homophily have high edge homophily ratio $h  \rightarrow 1$, while graphs with strong heterophily (i.e., low/weak homophily) have small edge homophily ratio $h \rightarrow 0$. 
\vspace{-0.15cm}
\end{definition}
The edge homophily ratio in Dfn.~\ref{dfn:homophily-ratio} gives an \textit{overall trend} for all the edges in the graph. The actual level of homophily may vary within different pairs of node classes, i.e., there is different tendency of connection between each pair of classes. In App.~\ref{app:homophily}, we give more details about capturing these more complex network characteristics via an empirical \textit{class compatibility matrix} $\mathbf{H}$, whose $i,j$-th entry is the fraction of outgoing edges to nodes in class $j$ among all outgoing edges from nodes in class $i$. 

\textit{Heterophily $\neq$ Heterogeneity.} 
We remark that heterophily, which we study in this work, is a distinct network concept from heterogeneity. 
Formally, a network is heterogeneous~\cite{SunH12} if it has at least two types of nodes and different relationships between them (e.g., knowledge graphs), and homogeneous if it has a single type of nodes (e.g., users) and a single type of edges (e.g., friendship). The type of nodes in heterogeneous graphs does \textit{not} necessarily match the class labels $y_v$, therefore both homogeneous and heterogeneous networks may have different levels of homophily. %

\section{Learning Over Networks with Heterophily} %
\label{sec:design}

\begin{wraptable}{r}{0.42\textwidth}
    \vspace{-0.45cm}
    \centering
    \caption{Example of a heterophily setting ($h=0.1$) where existing GNNs fail to generalize, and a typical homophily setting ($h=0.7$): mean accuracy and standard deviation over three %
    runs %
    (cf.\ App.~\ref{app:synthetic}).}
    \label{tab:design-mechanism-results}
    \vspace{-0.15cm}
    \resizebox{0.42\textwidth}{!}{
        \begin{tabular}{lcc}
        \toprule
               & {$\mathbf{\mathit{h} = 0.1}$}  & {$\mathbf{\mathit{h} = 0.7}$}        \\ \midrule
        GCN~\cite{kipf2016semi}         & $37.14 {\scriptstyle \pm 4.60}$  & $ {84.52 \scriptstyle \pm 0.54}$ \\
        GAT~\cite{velickovic2018graph}   & $33.11 {\scriptstyle \pm 1.20}$                      & $ {84.03 \scriptstyle \pm 0.97}$  \\ 
        GCN-Cheby~\cite{defferrard2016convolutional}  & $68.10 {\scriptstyle \pm 1.75}$         & $ {84.92 \scriptstyle \pm 1.03}$ \\ 
        GraphSAGE~\cite{hamilton2017inductive} & $72.89 {\scriptstyle \pm 2.42}$  & \cellcolor{gray!15}${85.06 \scriptstyle \pm 0.51}$ \\ 
        MixHop~\cite{MixHop}    & $58.93 {\scriptstyle \pm 2.84}$        & $ {84.43 \scriptstyle \pm 0.94}$  \\ \midrule 
        MLP                     & \cellcolor{gray!15}$74.85 {\scriptstyle \pm 0.76}$   & $ {71.72 \scriptstyle \pm 0.62}$   \\ \midrule
        \method \textbf{(ours)}          & {$\mathbf{76.87{\scriptstyle \pm 0.43}}$}            & {$\mathbf{88.28 {\scriptstyle \pm 0.66}}$} \\ 
        \bottomrule
        \end{tabular}
    }
    \vspace{-0.3cm}
\end{wraptable}
While many GNN models have been proposed, most of them are designed under the assumption of homophily, and are not capable of handling heterophily.  %
As a motivating example, Table \ref{tab:design-mechanism-results} shows the mean classification accuracy for several leading GNN models on our synthetic benchmark \texttt{syn-cora}, where we can control the homophily/heterophily level (see App.~\ref{app:synthetic} for details on the data and setup).  Here we consider two homophily ratios, $h = 0.1$ and $h=0.7$, one for high heterophily and one for high homophily. 
We observe that for heterophily ($h=0.1$) all existing methods fail to perform better than a Multilayer Perceptron (MLP) with 1 hidden layer, a graph-agnostic baseline that relies solely on the node features for classification (differences in accuracy of MLP for different $h$ are due to randomness). Especially, GCN~\cite{kipf2016semi} and GAT~\cite{velickovic2018graph} show up to 42\% worse performance than MLP, highlighting that methods that work well %
under high homophily ($h=0.7$) may not be appropriate for networks with low/medium homophily. %

Motivated by this limitation, in the following subsections, we discuss and theoretically justify a set of key design choices that, when appropriately incorporated in a GNN framework, can improve the performance in the challenging heterophily settings. Then, we present \method, a model that, thanks to these designs, adapts well to %
both homophily and heterophily (Table~\ref{tab:design-mechanism-results}, last row). %
In Section~\ref{sec:exp}, we provide a comprehensive empirical analysis on both synthetic and real data with varying homophily levels, and show that %
{the identified designs significantly improve the performance of GNNs (not limited to \method) by effectively leveraging the graph structure in challenging heterophily settings, while maintaining competitive performance in homophily.}

\subsection{Effective Designs for Networks with Heterophily} %
\label{sec:design-choices}

We have identified three key designs that---when appropriately integrated---can help improve the performance of GNN models in heterophily settings: (D1) \mechanismOneSmall; (D2) \mechanismTwoSmall ; and (D3) \mechanismThreeSmall. %
{While these designs have been utilized separately in some prior works~\cite{hamilton2017inductive, defferrard2016convolutional, MixHop, XuLTSKJ18-jkn}, we are the first to discuss their importance \textit{under heterophily} by providing novel theoretical justifications and an extensive empirical analysis on a variety of datasets.}

\subsubsection{(D1) \mechanismOne} The first design entails encoding each ego-embedding (i.e., a node's embedding) \textit{separately} from the aggregated embeddings of its neighbors, since they are likely to be dissimilar in heterophily settings. %
Formally, the representation (or hidden state vector) learned for each node $v$ at round $k$ is given as:

 \vspace{-0.45cm}
{\small 
\begin{equation}
    \V{r}^{(k)}_v = \texttt{\mytexthl{COMBINE}}\left(\V{r}^{(k-1)}_v, \; \texttt{AGGR}(\{\V{r}^{(k-1)}_u: u \in \mymathhl{\neighNoSelfLoop(v)}\})\right),
    \label{eq:design1}
\end{equation}
}
 \vspace{-0.45cm}
 
the neighborhood $\neighNoSelfLoop(v)$ does \textit{not} include $v$ (no self-loops), %
the \texttt{AGGR} function aggregates representations \textit{only} from the neighbors (in some way---e.g., average), and \texttt{AGGR} and \texttt{COMBINE} may be followed by a non-linear transformation.
For heterophily, after aggregating the neighbors' representations, the definition of \texttt{COMBINE} (akin to `skip connection' between layers) is critical: a simple way to combine %
the ego- and the aggregated neighbor-embeddings without `mixing' them is with concatenation {as in GraphSAGE~\cite{hamilton2017inductive}}---rather than averaging \textit{all} of them as in the  GCN model by~\citet{kipf2016semi}.

\noindent \textit{Intuition.} %
In heterophily settings, by definition (Dfn.~\ref{dfn:heterophily-nets}), the class label ${y}_v$ and original features $\V{x}_v$ of a node and those of its neighboring nodes $\{(y_u, \V{x}_u): u \in \neighNoSelfLoop(v)\}$ (esp. the direct neighbors $\neighNoSelfLoop_1(v)$) may be different. However, the typical GCN design that mixes the embeddings through an
average~\cite{kipf2016semi} or weighted average~\cite{velickovic2018graph} as the
\texttt{COMBINE} function %
results in final embeddings that are similar across neighboring nodes (especially within a community or cluster) \textit{for any set of original features}~\cite{Rossi2019FromCT}. 
While this may work well in the case of homophily, where neighbors likely belong to the same cluster and class, it poses severe challenges in the case of heterophily: it is not possible to distinguish neighbors from different classes based on the (similar) learned representations. Choosing a \texttt{COMBINE} function that separates the representations of each node $v$ and its neighbors $\neighNoSelfLoop(v)$ allows for more expressiveness, where the skipped or non-aggregated representations can evolve separately over multiple rounds of propagation %
without %
becoming prohibitively similar. %

\noindent \textit{Theoretical Justification.} We prove theoretically that, under some conditions, a GCN layer that co-embeds ego- and neighbor-features 
is less capable of generalizing to heterophily than a layer that embeds them separately. 
 We measure its generalization ability by its robustness to test/train data deviations. %
We give the proof of the theorem %
in App.~\ref{app:proof-thm1}. Though the theorem applies to specific conditions, our empirical analysis shows that it holds in more general cases (\S~\ref{sec:exp}).
\begin{theorem}
\label{thm:1}
Consider a %
graph $\graph$ without self-loops (\S~\ref{sec:preliminaries}) with node features $\V{x}_v = \mathrm{onehot}(y_v)$ for each node $v$, and an equal number of nodes per class $y\in\setY$ in the training set $\setT$.
Also assume that all nodes in $\setT$ have degree $d$, and proportion $h$ of their neighbors belong to the same class, while proportion $\tfrac{1-h}{|\setY|-1}$ of them belong to any other class (uniformly). 
Then for $h < \tfrac{1-|\setY|+2d}{2|\setY|d}$, a simple GCN layer formulated as $(\matA+\matI)\matX\matW$ %
is less robust, i.e.,  misclassifies a node for smaller 
train/test data deviations, than a $\matA\matX\matW$ layer %
that separates the ego- and neighbor-embeddings.
\end{theorem}

\noindent \textit{Observations.} In Table~\ref{tab:design-mechanism-results}, we observe that GCN, GAT, and MixHop, which `mix' the ego- and neighbor-embeddings explicitly\footnote{\label{mixhop-self-loops} 
These models consider self-loops, which turn each ego also into a neighbor, and thus mix the ego- and neighbor-representations. E.g., GCN and MixHop operate on the symmetric normalized adjacency matrix augmented with self-loops: 
$\matAh = \matDh^{-\frac{1}{2}} (\matA + \matI) \matDh^{-\frac{1}{2}},$ where $\matI$ is the identity and $\matDh$ the degree matrix of $\matA+\matI$.
}, 
perform poorly in the heterophily setting. On the other hand, GraphSAGE that separates the embeddings (e.g., it concatenates the two embeddings and then applies a non-linear transformation) achieves 33-40\% better performance in this setting. %

\subsubsection{(D2) \mechanismTwo} The second design involves explicitly aggregating information from higher-order neighborhoods in each round $k$, beyond the immediate neighbors of each node: %
{\small 
\begin{equation}
    \V{r}^{(k)}_v = \texttt{COMBINE}\left(\V{r}^{(k-1)}_v, \; \texttt{AGGR}(\{\V{r}^{(k-1)}_u: u \in \mymathhl{N_1(v)}\}), \; \texttt{AGGR}(\{\V{r}^{(k-1)}_u: u \in \mymathhl{N_2(v)}\}), \ldots \right)
\end{equation}
}%
where $N_i(v)$ denotes the neighbors of $v$ at \textit{exactly} $i$ hops away, and the \texttt{AGGR} functions applied to different neighborhoods can be the same or different. This design---employed in GCN-Cheby~\cite{defferrard2016convolutional} and MixHop~\cite{MixHop}---augments the \textit{implicit} aggregation over higher-order neighborhoods that most GNN models achieve through multiple rounds of first-order propagation based on variants of Eq.~\eqref{eq:design1}.

\noindent \textit{Intuition.} To show why higher-order neighborhoods 
help in the heterophily settings, we first define %
\textit{homophily-dominant} and \textit{heterophily-dominant} neighborhoods: %
\begin{definition}
$N(v)$ is expectedly homophily-dominant if $P(y_u = y_v|y_v) \geq P(y_u = y|y_v), \forall u\in N(v)$ and $y \in \setY \neq y_v$. If the opposite inequality holds, $N(v)$ is expectedly heterophily-dominant.
\end{definition}
From this definition, we can see that expectedly homophily-dominant neighborhoods are more beneficial for GNN layers, as %
in such neighborhoods the class label $y_v$ of each node $v$ can \textit{in expectation} be determined by the majority of the class labels in $N(v)$. 
In the case of heterophily, 
we have seen empirically that although the immediate neighborhoods may be heterophily-dominant, the higher-order neighborhoods may be homophily-dominant and thus provide more relevant context. 
{This observation is also confirmed by  recent works~\cite{altenburger2018monophily,chin2019decoupled} in the context of binary attribute prediction.}

\noindent \textit{Theoretical Justification.} Below we formalize the above observation for 2-hop neighborhoods {under non-binary attributes (labels)}, and prove one case when they are homophily-dominant in App.~\ref{app:proof-thm2}:

\begin{theorem}
\label{thm:2}
\vspace{-0.1cm}
Consider 
a %
graph $\graph$  without self-loops (\S~\ref{sec:preliminaries})  with label set $\setY$, 
where for each node $v$, its neighbors' class labels $\{y_u: u \in N(v)\}$ are conditionally independent given $y_v$, and $P(y_u=y_v|y_v) = h$, $P(y_u=y|y_v) = \frac{1-h}{|\setY|-1}, \forall y \neq y_v$. 
Then, the 2-hop neighborhood $N_2(v)$ for a node $v$ will always be homophily-dominant in expectation. %
\end{theorem}

\noindent \textit{Observations.} Under heterophily ($h=0.1$), GCN-Cheby, %
which models different neighborhoods by combining Chebyshev polynomials to approximate a higher-order graph convolution operation~\cite{defferrard2016convolutional}, outperforms GCN and GAT, which aggregate over only the immediate neighbors $N_1$, by up to +31\% (Table~\ref{tab:design-mechanism-results}). 
MixHop, which explicitly models 1-hop and 2-hop neighborhoods (though `mixes' the ego- and neighbor-embeddings$^\text{1}$, violating design D1), also outperforms these two models.  %

\subsubsection{(D3) \mechanismThree} The third design combines the intermediate representations of each node at the final layer: %
\vspace{-0.3cm}
{\small 
\begin{equation}
    \V{r}^{(\text{final})}_v = \texttt{COMBINE}\left(\mymathhl{\V{r}^{(1)}_v, \V{r}^{(2)}_v, \ldots,} \V{r}^{(K)}_v \right)
\end{equation}
}%
to explicitly capture local \textit{and} global information via \texttt{COMBINE} functions that leverage each representation separately--e.g., concatenation,  LSTM-attention~\cite{XuLTSKJ18-jkn}.
This design is 
introduced in jumping knowledge networks~\cite{XuLTSKJ18-jkn} and shown to increase the representation power of GCNs under \textit{homophily}.

\noindent \textit{Intuition.}  %
Intuitively, each round collects information 
with different locality---earlier rounds are more local, while later rounds capture increasingly more global information (implicitly, via propagation). Similar to D2 (which models explicit neighborhoods), this design models the distribution %
of neighbor representations %
in low-homophily networks more accurately.
It also allows the class prediction to leverage different neighborhood ranges %
in different networks, adapting to their structural properties.

\noindent \textit{Theoretical Justification.} The benefit of combining intermediate representations can be theoretically explained from the spectral perspective. %
Assuming a GCN-style layer---where propagation can be viewed as spectral filtering---, the {higher order polynomials of the normalized} adjacency matrix $\matA$ is a low-pass filter~\cite{wu2019simplifying}, so intermediate outputs from  earlier rounds contain higher-frequency components than outputs from later rounds. At the same time, the following theorem holds for graphs with heterophily, where we view class labels as graph signals (as in graph signal processing): 
\begin{theorem}
\label{thm:3}
\vspace{-0.2cm}
Consider %
graph signals (label vectors) $\V{s}, \V{t} \in \{0, 1\}^{|\vertexSet|}$ defined on an undirected graph $\graph$ with edge homophily ratios $h_s$ and $h_t$, respectively. %
If $h_s < h_t$, then signal $\V{s}$ has higher energy (Dfn.~\ref{def:app-laplacian-spectrum}) in high-frequency components than $\V{t}$ in the spectrum of unnormalized graph Laplacian $\matL$. 
\vspace{-0.2cm}
\end{theorem}
In other words, in heterophily settings, the label distribution contains more information at higher than lower frequencies (see proof in App.~\ref{app:proof-thm3}). %
Thus, by combining the intermediate outputs from different layers, this design captures both low- and high-frequency components in the final representation, which is critical in heterophily settings, and allows for more expressiveness in the general setting.

\noindent \textit{Observations.} %
By concatenating the intermediate representations from two rounds with the embedded ego-representation (following the jumping knowledge framework~\cite{XuLTSKJ18-jkn}), GCN's accuracy increases to  $58.93\% {\scriptstyle \pm3.17}$ for $h=0.1$, a 20\% improvement over its counterpart without design D3 (Table~\ref{tab:design-mechanism-results}).

\textbf{Summary of designs $\;$} To sum up, 
D1 models (at each layer) the ego- and neighbor-representations \textit{distinctly},
D2 leverages (at each layer) representations of neighbors at different distances \textit{distinctly}, and 
D3 leverages (at the final layer) the learned ego-representations at previous layers \textit{distinctly}.

\subsection{\method: A Framework for Networks with Homophily or Heterophily} %
\label{sec:method}

We now describe \method, which 
{exemplifies how effectively combining designs D1-D3 can help better adapt to the whole spectrum of low-to-high homophily, while avoiding interference with other designs}.  
It has three stages (Alg.~\ref{algo:method}, App. \ref{app:algo}): \stepone feature embedding, \steptwo neighborhood aggregation, and \stepthree classification.

The \textit{feature embedding stage} {\bf \stepone} %
uses a graph-agnostic dense layer to generate for each node $v$ %
the feature embedding {\small $\V{r}_v^{(0)}\in\mathbb{R}^p$} based on its ego-feature 
{\small $\V{x}_v$: 
$\V{r}_v^{{\scriptstyle (0)}}=%
\sigma(\V{x}_v \matW_e)$}, where $\sigma$ is an optional non-linear function, and {\small $\matW_e\in\mathbb{R}^{F\times p}$} is a learnable weight matrix.

In the \textit{neighborhood aggregation stage} {\bf \steptwo}, the generated embeddings %
are aggregated and repeatedly updated within the node's neighborhood for $K$ rounds. 
Following designs D1 and D2, the neighborhood $N(v)$ of our framework involves two sub-neighborhoods without the egos: the 1-hop graph neighbors $\neighNoSelfLoop_1(v)$ and the 2-hop neighbors $\neighNoSelfLoop_2(v)$, as shown in Fig.~\ref{fig:neighborhoods}: %
{\small
\begin{equation}
    \V{r}^{(k)}_v =  \texttt{COMBINE}\left(\texttt{AGGR}\{\V{r}^{(k-1)}_u: u \in \neighNoSelfLoop_1(v)\}, \texttt{AGGR}\{\V{r}^{(k-1)}_u: u \in \neighNoSelfLoop_2(v)\}\right).
   \label{eq:h2gcn-neighbor-combine}
\end{equation}
}
We set \texttt{COMBINE} as concatenation (as to not mix different neighborhood ranges), and \texttt{AGGR} as a degree-normalized average of the neighbor-embeddings in sub-neighborhood $\neighNoSelfLoop_i(v)$: %
 {\small
 \begin{equation}
    \V{r}_{v}^{(k)} = \left(  \V{r}_{v,1}^{(k)} \Vert \V{r}_{v,2}^{(k)} \right) 
    \quad \text{and} \quad
    \textstyle \V{r}_{v,i}^{(k)} = \texttt{AGGR}\{\V{r}^{(k-1)}_u: u \in \neighNoSelfLoop_i(v)\} = \sum_{u \in \neighNoSelfLoop_i(v)} \V{r}_u^{(k-1)} {d_{v,i}^{-\sfrac{1}{2}} d_{u,i}^{-\sfrac{1}{2}}},
\end{equation}
 }%
where $d_{v,i} = |\neighNoSelfLoop_i(v)|$ is the $i$-hop degree of node $v$ (i.e., number of nodes in its $i$-hop neighborhood). 
Unlike Eq.~\eqref{eq:design1}, here we do \textit{not} combine the ego-embedding of node $v$ with the neighbor-embeddings. We found that removing the usual nonlinear transformations %
per round, as in SGC~\cite{wu2019simplifying}, works better (App.~\ref{app:qualitative-comp}), in which case we only need to include the ego-embedding in the {final} representation.
By  design D3, each node's final representation combines all its intermediate representations: %

\vspace{-0.55cm}
{\small
\begin{equation}
    \label{eq:h2gcn-final-combine}
    \V{r}^{(\text{final})}_v =  \texttt{COMBINE}\left( \V{r}^{(0)}_v, \V{r}^{(1)}_v, \ldots, \V{r}^{(K)}_v\right), 
\end{equation}
}%
where we empirically find concatenation works better than max-pooling~\cite{XuLTSKJ18-jkn} as the \texttt{COMBINE} function.

In the \textit{classification stage} {\bf \stepthree}, the node is classified based on its final embedding $\V{r}_{v}^{(\text{final})}$:
{\small
\begin{equation}
    y_v = \arg \max \{\mathrm{softmax}(\V{r}_{v}^{(\text{final})}\matW_c ) \},
\end{equation}
}
where $\matW_c\in\mathbb{R}^{(2^{K+1}-1)p \times |\setY|}$ is a learnable weight matrix. We visualize our framework in App.~\ref{app:algo}. %

\paragraph{Time complexity} %
The feature embedding stage \stepone takes $\mathrm{O}(\text{nnz}(\matX)\,p)$, where $\text{nnz}(\matX)$ is the number of non-0s in feature matrix $\matX \in \mathbb{R}^{n \times F}$, and $p$ is the dimension of the feature embeddings.
The neighborhood aggregation stage \steptwo takes $\mathrm{O}\left(|\edgeSet|d_{\mathrm{max}}\right)$ to derive the 2-hop neighborhoods via sparse-matrix multiplications, where $d_{\mathrm{max}}$ is the maximum degree of all nodes, and $\mathrm{O}\left(2^K (|\edgeSet| + |\edgeSet_2|) p\right)$ for $K$ rounds of aggregation, where $|\edgeSet_2| = \frac{1}{2}\sum_{v\in \vertexSet} |\neighNoSelfLoop_2(v)|$. %
We give a detailed analysis in App.~\ref{app:algo}. %

\vspace{-0.15cm}
\section{Other Related Work}
\label{sec:related}

We discuss relevant work on GNNs here, and give other related work (e.g., classification under heterophily) in Appendix~\ref{app:related}.
Besides the models mentioned above, there are various comprehensive reviews describing previously proposed architectures~\cite{dlgsurvey_tkde20,mlgraphs_arxiv20,gcnreview_csn19}. %
Recent work has investigated GNN's ability to capture graph information, proposing diagnostic measurements based on feature smoothness and label smoothness~\cite{hou2020measuring} that may guide the learning process.  To capture more graph information, other works generalize graph convolution outside of immediate neighborhoods.  For example, apart from MixHop~\cite{MixHop} (cf.\ \S~\ref{sec:design-choices}), %
Graph Diffusion Convolution~\cite{klicpera_diffusion_2019} replaces the adjacency matrix with a sparsified version of a diffusion matrix (e.g., heat kernel or PageRank).  Geom-GCN~\cite{Pei2020Geom-GCN} precomputes unsupervised node embeddings and uses neighborhoods defined by geometric relationships in the resulting latent space to define graph convolution.  
Some of these works~\cite{MixHop,Pei2020Geom-GCN,hou2020measuring} 
acknowledge the challenges of learning from graphs with heterophily. %
Others have noted that node labels may have complex relationships that should be modeled directly. %
For instance, Graph Agreement Models~\cite{gam} augment the classification task with an agreement task, co-training a model to predict whether pairs of nodes %
share the same label; Graph Markov Neural Networks~\cite{qu2019gmnn} model the joint label distribution with a conditional random field, trained with expectation maximization using GNNs; %
{Correlated Graph Neural Networks~\cite{jia2020residual} model the correlation structure in the residuals of a regression task with a multivariate Gaussian, and can learn negative label correlations for neighbors in heterophily  (for binary class labels);}
{and the recent CPGNN~\cite{zhu2020graph} method models more complex label correlations by integrating the compatibility matrix notion from belief propagation~\cite{GatterbauerGKF15} into GNNs.}

\begin{wraptable}{r}{0.35\textwidth}
    \vspace{-0.5cm}
	\centering
	\caption{Design Comparison.}%
	\vspace{-0.2cm}
	\label{tab:comparison}
    {\scriptsize
	\begin{tabular}{lccc}
		\toprule 
		\textbf{Method} & \textbf{D1} & \textbf{D2} & \textbf{D3}   \\ 
		\midrule 
	GCN~\cite{kipf2016semi} & \xmark & \xmark & \xmark \\
	GAT~\cite{velickovic2018graph} & \xmark & \xmark & \xmark \\
    GCN-Cheby~\cite{defferrard2016convolutional} & \xmark & \cmark & \xmark \\
	GraphSAGE~\cite{hamilton2017inductive} & \cmark & \xmark & \xmark \\
	MixHop~\cite{MixHop} & \xmark & \cmark &  \xmark \\ \midrule
	\method (proposed) & \cmark & \cmark & \cmark \\ \bottomrule
	\end{tabular}
	}
	\vspace{-0.5cm}
\end{wraptable}
\textbf{Comparison of \method to existing GNN models} $\;$ %
As shown in Table~\ref{tab:comparison}, \method differs from existing GNN models with respect to designs D1-D3, and their implementations (we give more details in App.~\ref{app:algo}).  %
Notably, \method learns a graph-agnostic feature embedding in stage \textbf{(S1)}, and skips the non-linear embeddings of aggregated representations per round that other models use (e.g., GraphSAGE, MixHop, GCN), resulting in a simpler yet powerful architecture.

\vspace{-0.15cm}
\section{Empirical Evaluation}
\label{sec:exp}

We %
{show the significance of designs D1-D3 on synthetic and real graphs with low-to-high homophily (Tab.~\ref{tab:5-synthetic-data}, \ref{tab:5-real-results}) via an ablation study of \method and comparison of models with and without the designs.}

\textbf{Baseline models $\;$} 
We consider MLP with 1 hidden layer, and all the methods listed in Table~\ref{tab:comparison}.
For \method, we model the first- and second-order neighborhoods ($\neighNoSelfLoop_1$ and $\neighNoSelfLoop_2$), and consider two variants: \method-1 uses one embedding round ($K=1$) and \method-2 uses two rounds ($K=2$). 
We tune all the models on the same train/validation splits (see App.~\ref{app:tuning} for details).

\vspace{-0.15cm}
\subsection{Evaluation on Synthetic Benchmarks}
\label{sec:eval-synthetic}

\begin{wrapfigure}{r}{0.36\textwidth}
    \vspace{-0.95cm}
    \centering
    \begin{subfigure}{0.35\textwidth}
        \centering
        \includegraphics[keepaspectratio, width=\textwidth,trim={0cm 0cm 2.0cm 2.6cm},clip]{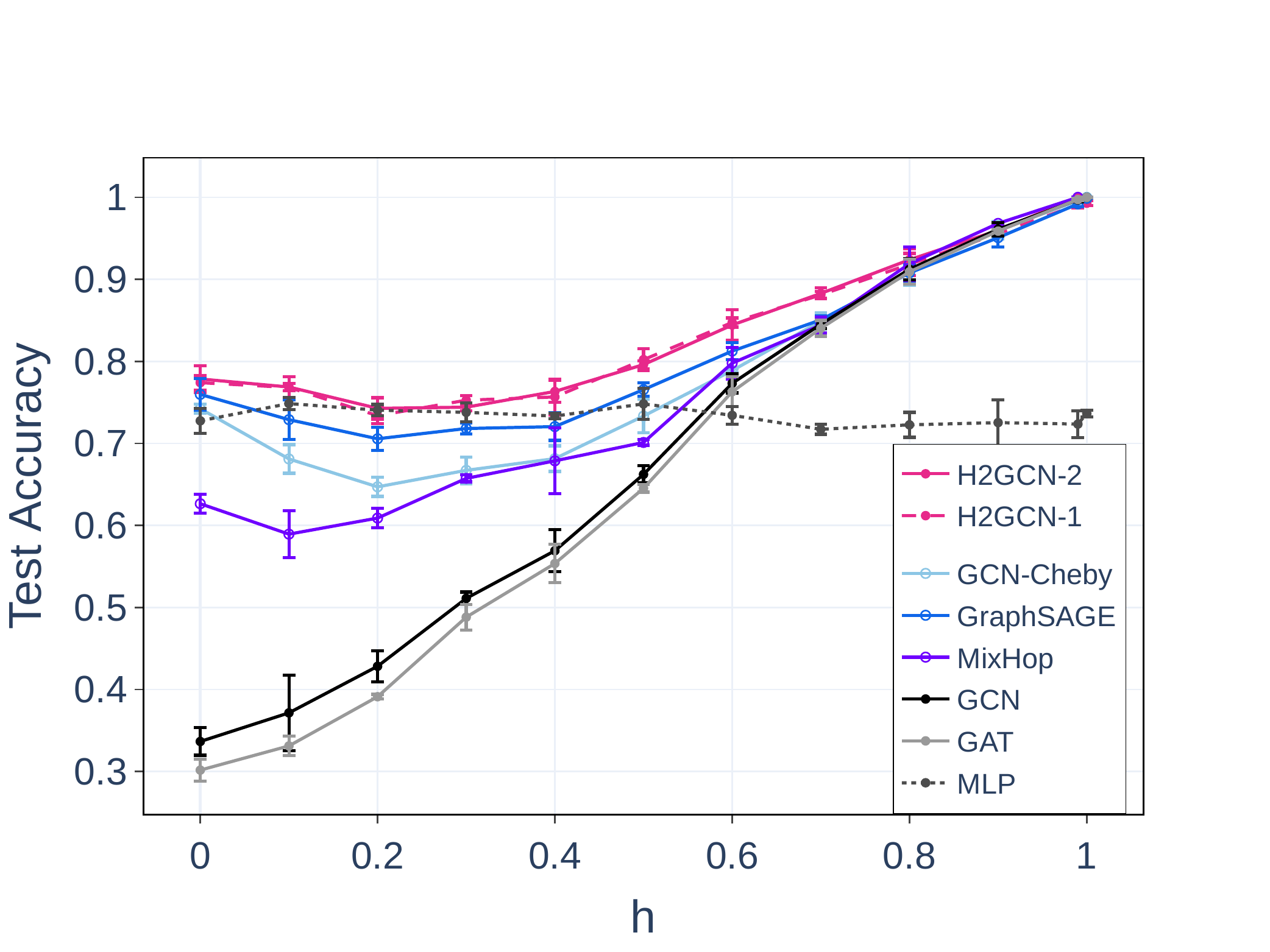}
        \vspace{-0.6cm}
        \caption{\texttt{syn-cora} (Table \ref{tab:app-syn-cora-results})}
        \label{fig:syn-cora}
    \end{subfigure}
    \begin{subfigure}{0.35\textwidth}
        \centering
        \includegraphics[keepaspectratio, width=\textwidth,trim={0cm 0cm 2.0cm 2.6cm},clip]{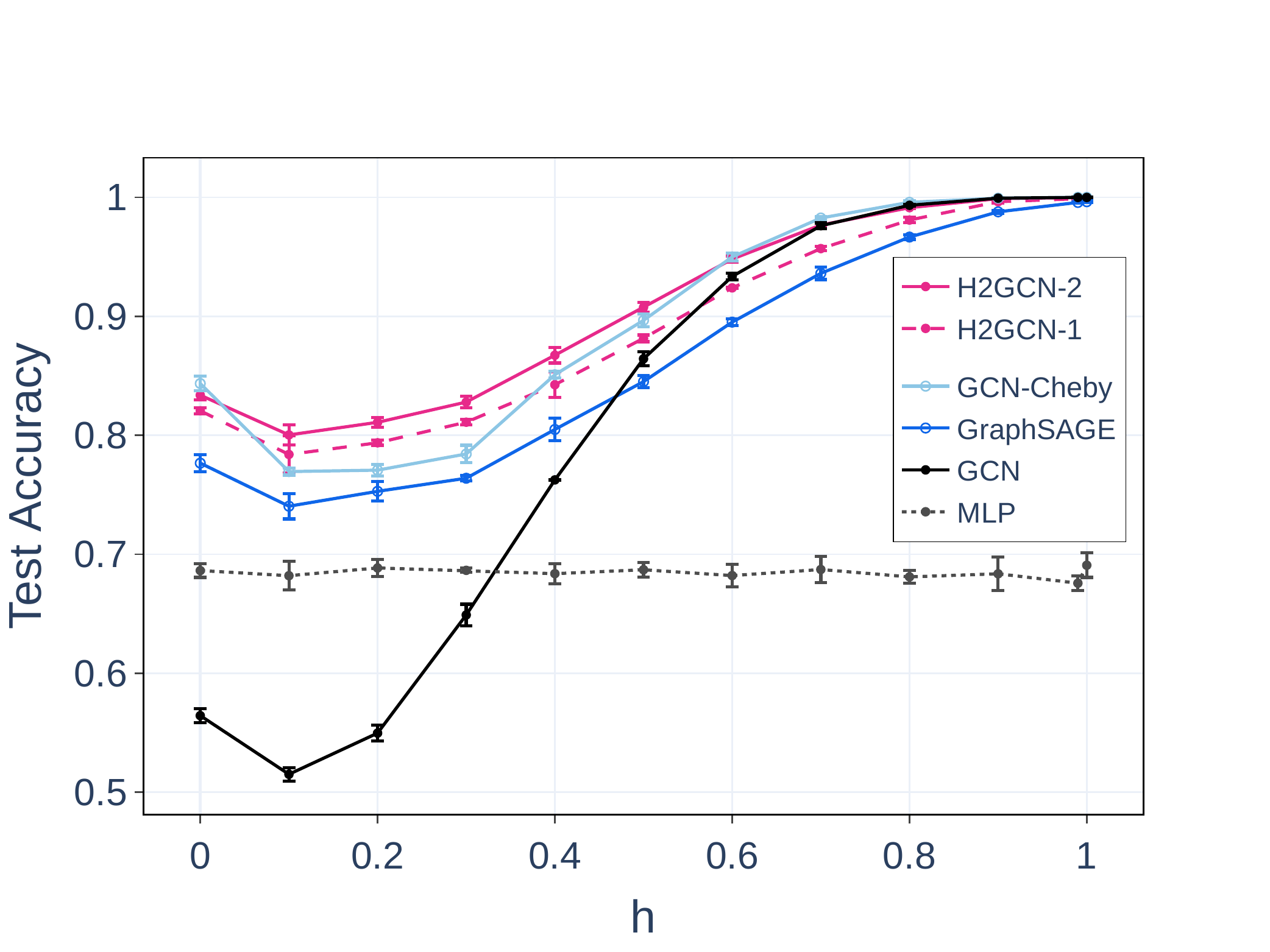}
        \vspace{-0.6cm}
        \caption{\texttt{syn-products} (Table \ref{tab:app-syn-products-results}). MixHop acc $<30\%$; GAT acc $<50\%$ for $h<0.4$.} %
        \label{fig:syn-products}
    \end{subfigure}
    \vspace{-0.1cm}
    \caption{Performance of GNN models on synthetic datasets. \method-2 outperforms baseline models in most heterophily settings, while tying with other models in homophily. %
    }
    \label{fig:5-syn-results}
    \vspace{-0.8cm}
\end{wrapfigure}
\paragraph{Synthetic datasets \& setup} We generate synthetic graphs with various homophily ratios $h$ (Tab.~\ref{tab:5-synthetic-data}) by adopting an approach similar to \cite{karimi2017visibility17}. %
In App.~\ref{app:synthetic}, we describe the data generation process, the experimental setup, and the data statistics in detail. All methods share the same training, validation and test splits (25\%, 25\%, 50\% per class), and we report the average accuracy and standard deviation (stdev) %
over three generated graphs %
per heterophily level and benchmark dataset.

\begin{table}[t!]
	\centering
	\vspace{-0.2cm}
	\caption{Statistics for Synthetic Datasets} %
	\label{tab:5-synthetic-data}
	{\scriptsize
	\begin{tabular}{lrrccccc}
		\toprule 
		\textbf{Benchmark Name} & \multicolumn{1}{c}{\textbf{\#Nodes $|\vertexSet|$}}  & \multicolumn{1}{c}{\textbf{\#Edges $|\edgeSet|$}} & \multicolumn{1}{c}{\textbf{\#Classes $|\setY|$}} & %
		\multicolumn{1}{c}{\textbf{\#Features $F$}} & \multicolumn{1}{c}{\textbf{Homophily $h$}} & \multicolumn{1}{c}{\textbf{\#Graphs}} \\ \midrule %
	\texttt{syn-cora} & $1,490$ & $2,965$ to $2,968$ & 5 & %
	\texttt{cora} \cite{sen2008collective, yang2016revisiting} & [0, 0.1, \ldots, 1] & $ 33 $ (3 per $h$)\\
	\texttt{syn-products} & $10,000$ & $59,640$ to $59,648$ & 10 & %
	\texttt{ogbn-products}~\cite{hu2020ogb} & [0, 0.1, \ldots, 1] & $ 33 $ (3 per $h$)\\ \bottomrule
	\end{tabular}
	}
	\vspace{-0.5cm}
\end{table}

\textbf{Model comparison $\;$} %
Figure~\ref{fig:5-syn-results} shows the mean test accuracy (and stdev) over all random splits of our synthetic benchmarks. 
We observe similar trends on both benchmarks: 
\method has the best trend overall, outperforming the baseline models in most heterophily settings, while tying with other models in homophily.
The performance of GCN, GAT and MixHop, which mix the ego- and neighbor-embeddings, increases with respect to the homophily level.  
But, while they achieve near-perfect accuracy under strong homophily ($h \rightarrow 1$), they are significantly less accurate than MLP (near-flat performance curve as it is graph-agnostic) for many heterophily settings. GraphSAGE and GCN-Cheby, which leverage some of the identified designs D1-D3 (Table~\ref{tab:comparison}, \S~\ref{sec:design}), are more competitive in such settings. %
We note that all the methods---except GCN and GAT---learn more effectively under perfect heterophily ($h$=$0$) than weaker settings (e.g., $h\in[0.1,0.3]$), as evidenced by the J-shaped performance curves in low-homophily ranges.

\begin{figure}[t!]
    \centering
    \begin{subfigure}{0.235\textwidth}
        \centering
        \includegraphics[keepaspectratio, width=\textwidth,trim={0 0 2.2cm 2.6cm},clip]{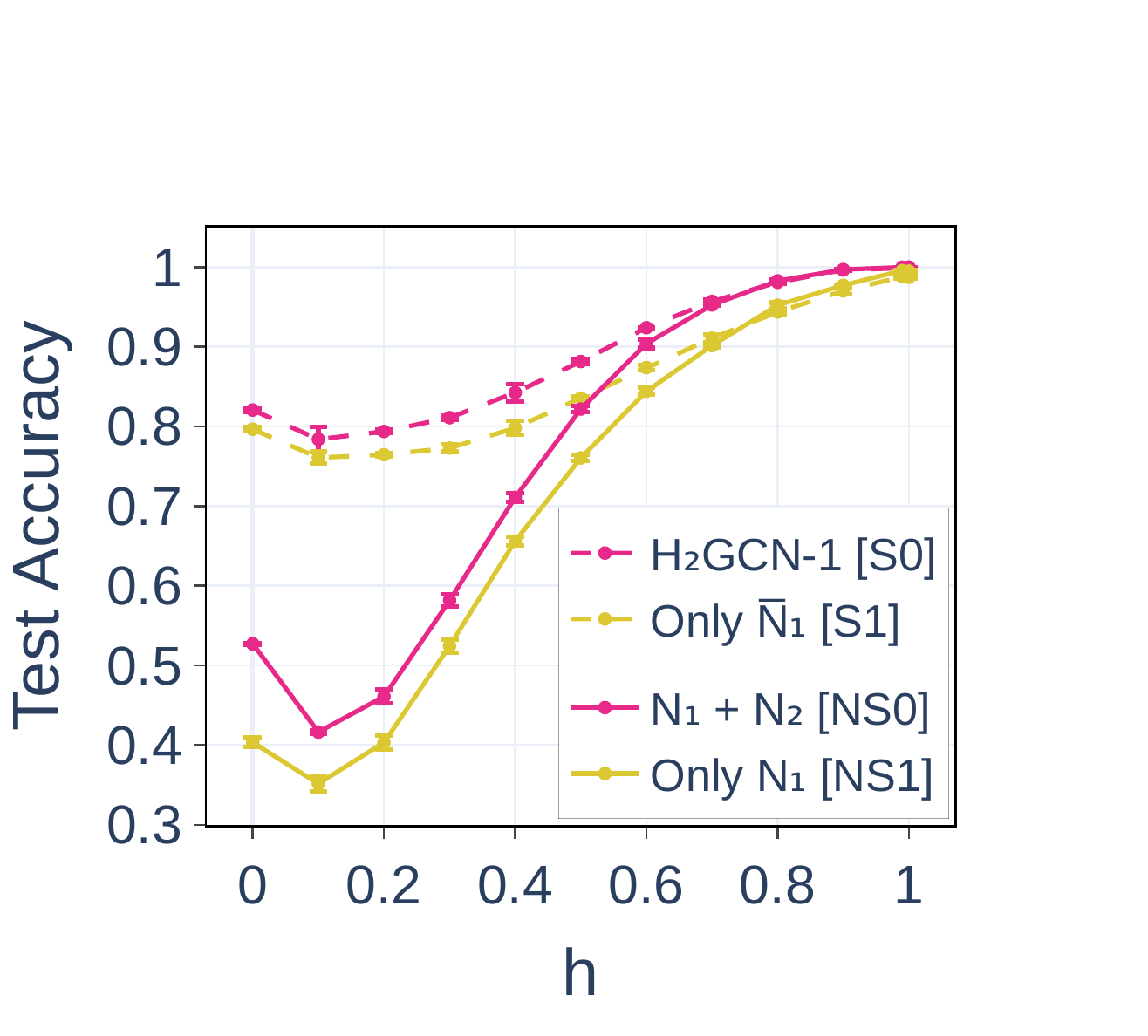}
        \caption{Design D1: Embedding separation.}%
        \label{fig:5-ablation-1}
    \end{subfigure}
    ~
    \begin{subfigure}{0.235\textwidth}
        \centering
        \includegraphics[keepaspectratio, width=\textwidth,trim={0 0 2.2cm 2.6cm},clip]{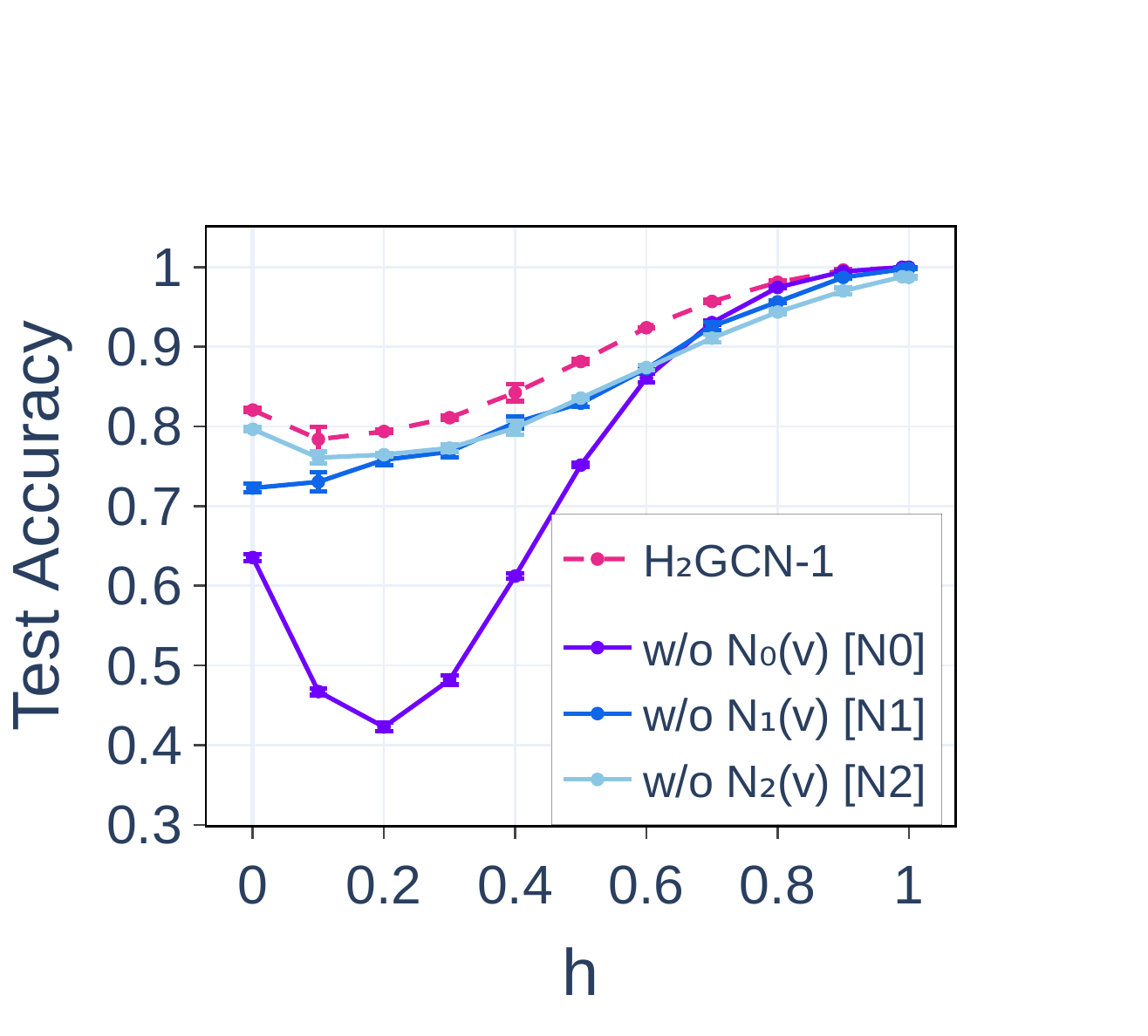}
        \caption{Design D2: Higher-order neighborhoods.}
        \label{fig:5-ablation-2}
    \end{subfigure}
    ~
    \begin{subfigure}{0.235\textwidth}
        \centering
        \includegraphics[keepaspectratio, width=\textwidth,trim={0 0 2.2cm 2.6cm},clip]{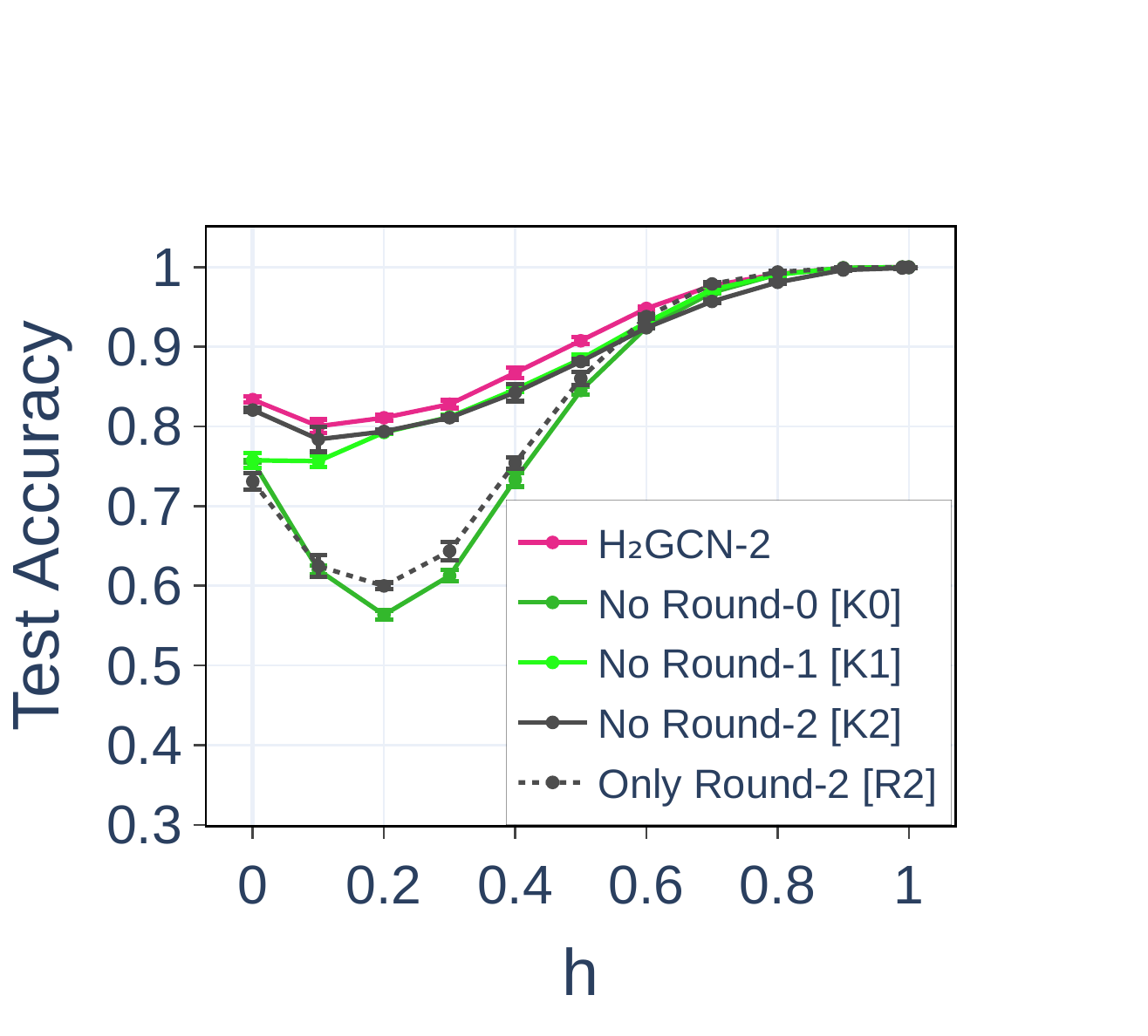}
        \caption{Design D3: Intermediate representations.} %
        \label{fig:5-ablation-3}
    \end{subfigure}
    ~
    \begin{minipage}{0.24\textwidth}
        \centering
    \begin{subfigure}{\textwidth}
        \centering
        \includegraphics[keepaspectratio, width=\textwidth]{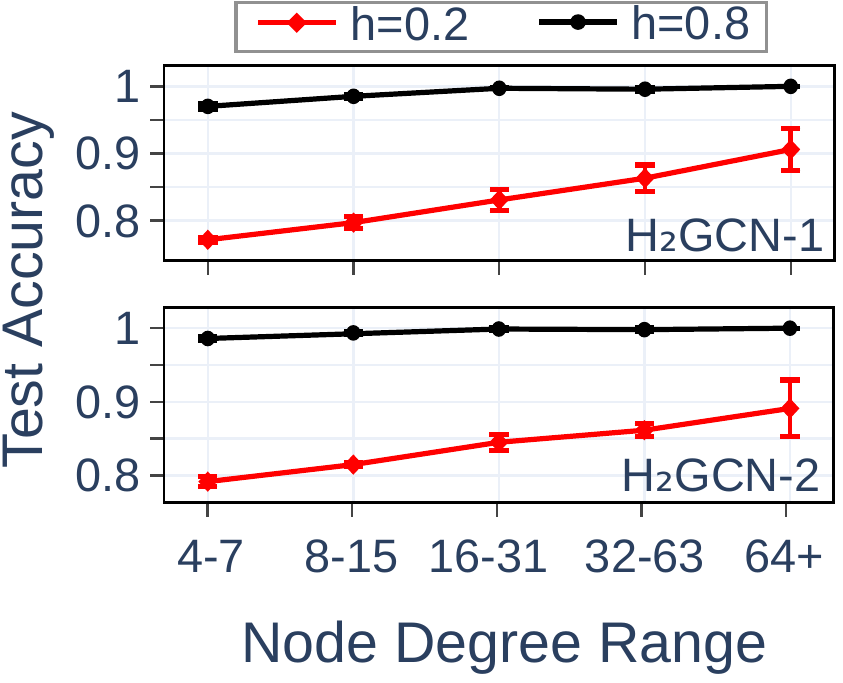} %
        \vspace{-0.25cm}
        \caption{Accuracy per degree in hetero/homo-phily.}
        \label{fig:5-degree-2}
    \end{subfigure}
    \end{minipage}
    \vspace{-0.2cm}
    \caption{(\subref{fig:5-ablation-1})-(\subref{fig:5-ablation-3}): Significance of design choices D1-D3 via ablation studies. %
    (\subref{fig:5-degree-2}): Performance of \method for different node degree ranges. 
    In heterophily, the performance gap between low- and high-degree nodes is significantly larger than in homophily, i.e., low-degree nodes pose challenges.
    }
    \label{fig:design-ablations}
    \vspace{-0.5cm}
\end{figure}

\textbf{Significance of design choices $\;$}
Using \texttt{syn-products}, we show the significance of designs D1-D3 (\S~\ref{sec:design-choices}) through ablation studies with variants of \method (Fig.~\ref{fig:design-ablations}, Table~\ref{tab:design-ablations}).

\textit{\emph{(D1)} \mechanismOne.} 
We consider \method-1 variants that \textit{separate} the ego- and neighbor-embeddings and model: 
(\texttt{S0}) neighborhoods $\neighNoSelfLoop_1$ and $\neighNoSelfLoop_2$ (i.e., \method-1); 
(\texttt{S1}) only the 1-hop neighborhood $\neighNoSelfLoop_1$ in Eq.~\eqref{eq:h2gcn-neighbor-combine};  
and their counterparts that do \textit{not separate} the two embeddings and use:
(\texttt{NS0}) neighborhoods $N_1$ and $N_2$ (including $v$); and
(\texttt{NS1}) only the 1-hop neighborhood $N_1$. 
Figure~\ref{fig:5-ablation-1} shows that the  variants that learn separate embedding functions significantly outperform the others (\texttt{NS0/1})
in heterophily settings ($h<0.7$) by up to $40\%$, which shows that design D1 is critical for success in heterophily.
 \method-1 (\texttt{S0}) performs best
in homophily.

\textit{\emph{(D2)} \mechanismTwo.} 
For this design, we consider three variants of \method-1 without specific neighborhoods: 
(\texttt{N0}) without the 0-hop neighborhood $N_0(v)=v$ (i.e, the ego-embedding) %
(\texttt{N1}) without $\neighNoSelfLoop_1(v)$; and 
(\texttt{N2}) without $\neighNoSelfLoop_2(v).$ 
Figure~\ref{fig:5-ablation-2} shows that \method-1 consistently performs better than all the variants, indicating that combining all sub-neighborhoods works best.
Among the variants, in heterophily settings, $N_0(v)$ contributes most to the performance (\texttt{N0} causes significant decrease in accuracy), followed by $\neighNoSelfLoop_1(v)$, and $\neighNoSelfLoop_2(v)$. However, when $h \geq 0.7$, the importance of sub-neighborhoods is reversed. %
Thus, the ego-features are the most important in heterophily, and higher-order neighborhoods contribute the most in homophily. The design of \method allows it to effectively combine information from different neighborhoods, adapting to all levels of homophily. %

\textit{\emph{(D3)} \mechanismThree.} %
We consider three variants (\texttt{K}-\texttt{0},\texttt{1},\texttt{2}) of \method-2 that drop from the final representation of Eq.~\eqref{eq:h2gcn-final-combine} the $0^{th}$, $1^{st}$ or $2^{nd}$-round intermediate representation, respectively. We also consider only the $2^{nd}$ intermediate representation as final, 
which is akin to what the other GNN models do. Figure~\ref{fig:5-ablation-3} shows that \method-2, which combines all the intermediate representations, performs the best, followed by the variant \texttt{K2} that skips the round-2 representation. The ego-embedding %
is the most important for heterophily $h\leq0.5$ (see trend of \texttt{K0}).

\textbf{The challenging case of low-degree nodes $\;$} 
Figure~\ref{fig:5-degree-2} plots the mean accuracy of \method variants on \texttt{syn-products} for different node degree ranges  both in a heterophily and a homophily setting ($h\in\{0.2,0.8\}$). 
We observe that under heterophily there is a significantly bigger performance gap between low- and high-degree nodes: 13\% for \method-1 (10\% for \method-2) vs.\ less than 3\% under homophily. 
This is likely due to the importance of the  \textit{distribution} of class labels in each neighborhood under heterophily, which is harder to estimate accurately for low-degree nodes with few neighbors. On the other hand, in homophily, neighbors are likely to have similar classes $y\in\setY$, so the neighborhood size does not have as significant impact on the accuracy.

\vspace{-0.2cm}
\subsection{Evaluation on Real Benchmarks}
\label{sec:real-eval}
\vspace{-0.2cm}
\begin{wraptable}{r}{0.45\textwidth}
    \vspace{-1cm}
	\centering
	\caption{Real benchmarks: 
	Average rank per method (and their employed designs among D1-D3) under heterophily (benchmarks with $h\leq 0.3$),
	homophily ($h \geq 0.7$), and across the full spectrum (``Overall'').
	The ``*'' denotes ranks based on results reported in~\cite{Pei2020Geom-GCN}.
	}
	\label{tab:5-real-rank}
    {\scriptsize
	\begin{tabular}{lccHc}
		\toprule 
		\textbf{Method (Designs)} & \textbf{Het.} & \textbf{Hom.} & \textbf{Overall} & \textbf{Overall}  \\
		\midrule 
	\textbf{\method-1} (D1, D2, D3) & 3.8 & 3.0 & 3.5 & 3.6\\
	\textbf{\method-2} (D1, D2, D3) & 4.0 & 2.0 & 3.1 & 3.3 \\
    \textbf{GraphSAGE} (D1) & 5.0 & 6.0 & 5.7 & 5.3 \\
	\textbf{GCN-Cheby} (D2) & 7.0 & 6.3 & 6.5 & 6.8 \\
	\textbf{MixHop} (D2) & 6.5 & 6.0 & 6.4 & 6.3 \\ 
	\midrule
	\textbf{GraphSAGE+JK} (D1, D3) & 5.0 & 7.0 & 5.9 & 5.7 \\
	\textbf{GCN-Cheby+JK} (D2, D3) & 3.7 & 7.7 & 5.0 & 5.0 \\
	\textbf{GCN+JK} (D3) & 7.2 & 8.7 & 7.5 & 7.7 \\
	\midrule
	\textbf{GCN} & 9.8 & 5.3 & 7.7 & 8.3 \\
	\textbf{GAT} & 11.5 & 10.7 & 11.1 & 11.2 \\
	\textbf{GEOM-GCN}* & 8.2 & 4.0 & 6.8 & 6.8 \\
	\midrule
	\textbf{MLP} & 6.2 & 11.3 & 8.2 & 7.9 \\
	\bottomrule
	\end{tabular}
	}
	\vspace{-0.5cm}
\end{wraptable}
\paragraph{Real datasets \& setup} We now evaluate the performance of our model and existing GNNs on a variety of real-world datasets~\cite{tang2009social-fc,rozemberczki2019multiscale,sen2008collective, namata2012query,bojchevski2018deep, shchur2018pitfalls} %
with %
edge homophily ratio $h$ ranging from strong heterophily to strong homophily, going beyond the traditional Cora, Pubmed and Citeseer graphs that have strong homophily (hence the good performance of existing GNNs on them). We summarize the data in Table~\ref{tab:5-real-results}, and describe them in App.~\ref{app:real}, where we also point out potential data limitations. For all benchmarks (except \texttt{Cora-Full}), we use the %
feature vectors, class labels, and 10 random splits (48\%/32\%/20\% of nodes per class for train/validation/test\footnote{\cite{Pei2020Geom-GCN} claims that the ratios %
are 60\%/20\%/20\%, which is different from the actual data splits shared on GitHub.}) provided by \cite{Pei2020Geom-GCN}. 
For Cora-Full, we generate 3 random splits, with 25\%/25\%/50\% of nodes per class for train/validation/test. %

\setlength{\tabcolsep}{3pt}
\begin{table}%
    \centering
    \vspace{-0.35cm}
    \caption{%
    Real data: mean accuracy $\pm$ stdev over different data splits. Best model per benchmark highlighted in gray. The ``*'' results are obtained from~\cite{Pei2020Geom-GCN} and ``N/A'' denotes non-reported results. %
    }
    \label{tab:5-real-results}
    \begin{adjustbox}{width=\textwidth}
    \begin{tabular}{lccccccc  cccH} %
    \toprule
       &  \texttt{\bf Texas}           &   \texttt{\bf Wisconsin}           &   \texttt{\bf Actor}            &   \texttt{\bf Squirrel}   &   \texttt{\bf Chameleon} & \texttt{\bf Cornell}        &  \texttt{\bf Cora Full}   &   \texttt{\bf Citeseer}           &   \texttt{\bf Pubmed}            &   \texttt{\bf Cora} &  \\
          \textbf{Hom.\ ratio} $h$ & \textbf{0.11} & \textbf{0.21} & \textbf{0.22} & \textbf{0.22} & \textbf{0.23} & \textbf{0.3} & \textbf{0.57} & \textbf{0.74} & \textbf{0.8} & \textbf{0.81} & {\multirow{4}{*}{\rotatebox[origin=c]{90}{\textbf{Avg Rank}}}}\\
		\textbf{\#Nodes $|\vertexSet|$} & 183 & 251 & 7,600 & 5,201 & 2,277 & 183 & 19,793 & 3,327 & 19,717 & 2,708 \\
		\textbf{\#Edges $|\edgeSet|$} & 295 & 466 & 26,752 & 198,493 & 31,421 & 280 & 63,421 & 4,676 & 44,327 & 5,278 & \\
		\textbf{\#Classes $|\setY|$} & 5 & 5 & 5 & 5 & 5 & 5 & 70 & 7 & 3 & 6 & \\ 
    \midrule
    \midrule
       {\method-1} & \cellcolor{gray!15}$84.86{\scriptstyle\pm6.77}$ & \cellcolor{gray!15}$86.67{\scriptstyle\pm4.69}$ & \cellcolor{gray!15}$35.86{\scriptstyle\pm1.03}$ & $36.42{\scriptstyle\pm1.89}$ & $57.11{\scriptstyle\pm1.58}$ & \cellcolor{gray!15}$82.16{\scriptstyle\pm4.80}$ & $68.13{\scriptstyle\pm0.49}$ & $77.07{\scriptstyle\pm1.64}$ & $89.40{\scriptstyle\pm0.34}$ & $86.92{\scriptstyle\pm1.37}$ & 3.5\\
       {\method-2} & $82.16{\scriptstyle\pm5.28}$ & $85.88{\scriptstyle\pm4.22}$ & $35.62{\scriptstyle\pm1.30}$ & $37.90{\scriptstyle\pm2.02}$ & $59.39{\scriptstyle\pm1.98}$ & \cellcolor{gray!15}$82.16{\scriptstyle\pm6.00}$ & \cellcolor{gray!15}$69.05{\scriptstyle\pm0.37}$ & $76.88{\scriptstyle\pm1.77}$ & $89.59{\scriptstyle\pm0.33}$ & \cellcolor{gray!15}$87.81{\scriptstyle\pm1.35}$ & 3.1\\
       {GraphSAGE} & $82.43{\scriptstyle\pm6.14}$ & $81.18{\scriptstyle\pm5.56}$ & $34.23{\scriptstyle\pm0.99}$ & $41.61{\scriptstyle\pm0.74}$ & $58.73{\scriptstyle\pm1.68}$ & $75.95{\scriptstyle\pm5.01}$ & $65.14{\scriptstyle\pm0.75}$ & $76.04{\scriptstyle\pm1.30}$ & $88.45{\scriptstyle\pm0.50}$ & $86.90{\scriptstyle\pm1.04}$ & 5.7\\
       {GCN-Cheby} & $77.30{\scriptstyle\pm4.07}$ & $79.41{\scriptstyle\pm4.46}$ & $34.11{\scriptstyle\pm1.09}$ & $43.86{\scriptstyle\pm1.64}$ & $55.24{\scriptstyle\pm2.76}$ & $74.32{\scriptstyle\pm7.46}$ & $67.41{\scriptstyle\pm0.69}$ & $75.82{\scriptstyle\pm1.53}$ & $88.72{\scriptstyle\pm0.55}$ & $86.76{\scriptstyle\pm0.95}$ & 6.5 \\
	   {MixHop} & $77.84{\scriptstyle\pm7.73}$ & $75.88{\scriptstyle\pm4.90}$ & $32.22{\scriptstyle\pm2.34}$ & $43.80{\scriptstyle\pm1.48}$ & $60.50{\scriptstyle\pm2.53}$ & $73.51{\scriptstyle\pm6.34}$ & $65.59{\scriptstyle\pm0.34}$ & $76.26{\scriptstyle\pm1.33}$ & $85.31{\scriptstyle\pm0.61}$ & $87.61{\scriptstyle\pm0.85}$ & 6.4 \\ 
        \midrule
        {GraphSAGE+JK} & $83.78{\scriptstyle\pm2.21}$ & $81.96{\scriptstyle\pm4.96}$ & $34.28{\scriptstyle\pm1.01}$ & $40.85{\scriptstyle\pm1.29}$ & $58.11{\scriptstyle\pm1.97}$ & $75.68{\scriptstyle\pm4.03}$ & $65.31{\scriptstyle\pm0.58}$ & $76.05{\scriptstyle\pm1.37}$ & $88.34{\scriptstyle\pm0.62}$ & $85.96{\scriptstyle\pm0.83}$ & 5.9\\
        {Cheby+JK} & $78.38{\scriptstyle\pm6.37}$ & $82.55{\scriptstyle\pm4.57}$ & $35.14{\scriptstyle\pm1.37}$ & \cellcolor{gray!15}$45.03{\scriptstyle\pm1.73}$ & \cellcolor{gray!15}$63.79{\scriptstyle\pm2.27}$ & $74.59{\scriptstyle\pm7.87}$ & $66.87{\scriptstyle\pm0.29}$ & $74.98{\scriptstyle\pm1.18}$ & $89.07{\scriptstyle\pm0.30}$ & $85.49{\scriptstyle\pm1.27}$ & 5.0 \\
        {GCN+JK} & $66.49{\scriptstyle\pm6.64}$ & $74.31{\scriptstyle\pm6.43}$ & $34.18{\scriptstyle\pm0.85}$ & $40.45{\scriptstyle\pm1.61}$ & $63.42{\scriptstyle\pm2.00}$ & $64.59{\scriptstyle\pm8.68}$ & $66.72{\scriptstyle\pm0.61}$ & $74.51{\scriptstyle\pm1.75}$ & $88.41{\scriptstyle\pm0.45}$ & $85.79{\scriptstyle\pm0.92}$ & 7.5\\
       \midrule
	   {GCN} & $59.46{\scriptstyle\pm5.25}$ & $59.80{\scriptstyle\pm6.99}$ & $30.26{\scriptstyle\pm0.79}$ & $36.89{\scriptstyle\pm1.34}$ & $59.82{\scriptstyle\pm2.58}$ & $57.03{\scriptstyle\pm4.67}$ & $68.39{\scriptstyle\pm0.32}$ & $76.68{\scriptstyle\pm1.64}$ & $87.38{\scriptstyle\pm0.66}$ & $87.28{\scriptstyle\pm1.26}$ & 7.7 \\
	   {GAT} & $58.38{\scriptstyle\pm4.45}$ & $55.29{\scriptstyle\pm8.71}$ & $26.28{\scriptstyle\pm1.73}$ & $30.62{\scriptstyle\pm2.11}$ & $54.69{\scriptstyle\pm1.95}$ & $58.92{\scriptstyle\pm3.32}$ & $59.81{\scriptstyle\pm0.92}$ & $75.46{\scriptstyle\pm1.72}$ & $84.68{\scriptstyle\pm0.44}$ & $82.68{\scriptstyle\pm1.80}$ & 11.1 \\
	   {GEOM-GCN*} & $67.57$ & $64.12$ & $31.63$ & $38.14$ & $60.90$ & $60.81$ 
	   & N/A   & \cellcolor{gray!15}{$77.99$} &  \cellcolor{gray!15}{$90.05$} & $85.27$ & 6.8 \\
	   \midrule       
	   {MLP} & $81.89{\scriptstyle\pm4.78}$ & $85.29{\scriptstyle\pm3.61}$ & $35.76{\scriptstyle\pm0.98}$ & $29.68{\scriptstyle\pm1.81}$ & $46.36{\scriptstyle\pm2.52}$ & $81.08{\scriptstyle\pm6.37}$ & $58.76{\scriptstyle\pm0.50}$ & $72.41{\scriptstyle\pm2.18}$ & $86.65{\scriptstyle\pm0.35}$ & $74.75{\scriptstyle\pm2.22}$ & 8.2 \\
	   \bottomrule
    \end{tabular}
    \end{adjustbox}
    \vspace{-0.5cm}
\end{table}

\vspace{-0.25cm}
\paragraph{Effectiveness of design choices} Table~\ref{tab:5-real-rank} gives the average ranks of our \method variants and other models on real benchmarks with heterophily, homophily, and across the full spectrum. 
Table~\ref{tab:5-real-results} gives detailed results (mean accuracy and stdev) per benchmark. 
We observe that models which utilize all or subsets of our identified designs D1-D3 (\S~\ref{sec:design-choices}) perform significantly better than GCN and GAT which lack these designs, especially in heterophily.
Next, we discuss the effectiveness of each design.

(D1) \textit{\mechanismOne}. We compare GraphSAGE, which \textit{separates} the ego- and neighbor-embeddings, and GCN that does not. 
In heterophily settings, GraphSAGE has an average rank of 5.0 compared to 9.8 for GCN, and outperforms GCN in almost all heterophily benchmarks by up to 23\%. In homophily settings ($h \geq 0.7$), GraphSAGE ranks close to GCN (6.0 vs. 5.3), and GCN never outperforms GraphSAGE by more than 1\% in mean accuracy. These results support the importance of D1 for success in heterophily and comparable performance in homophily.

(D2) \textit{\mechanismTwo}. {To show the benefits of design D2 under heterophily, we compare the performance of GCN-Cheby and MixHop---which define higher-order graph convolutions---to that of (first-order) GCN. Under heterophily, 
GCN-Cheby (rank 7.0) and MixHop (rank 6.5) have better performance than GCN (rank 9.8), and outperform the latter in all but one heterophily benchmarks by up to 20\%. 
In most homophily benchmarks, the performance difference between these methods is less than 1\%. Our observations highlight the importance of D2, especially in heterophily. 
}

{(D3) \textit{\mechanismThree}. We compare GraphSAGE, GCN-Cheby and GCN to their corresponding variants enhanced with JK connections~\cite{XuLTSKJ18-jkn}. %
GCN and GCN-Cheby benefit significantly from D3 in heterophily: their average ranks improve (9.8 vs. 7.2 and 7 vs 3.7, respectively) and their mean accuracies increase by up to 14\% and 8\%, respectively, in heterophily benchmarks. Though GraphSAGE+JK performs better than GraphSAGE on half of the heterophily benchmarks, its average rank remains unchanged. 
This may be due to the marginal benefit of D3 when combined with D1, which GraphSAGE employs.
Under homophily, the performance with and without JK connections is similar (gaps mostly less than 2\%), %
matching the observations in \cite{XuLTSKJ18-jkn}.}

{While other design choices and implementation details may confound a comparative evaluation of D1-D3 in different models (motivating our introduction of \method and our ablation study in \S~\ref{sec:design-choices}), these observations support the effectiveness of our identified designs on diverse GNN architectures and real-world datasets, and affirm our findings in the ablation study.
We also observe that our \method variants, which combine the three identified designs, have consistently strong performance across the \textit{full spectrum} of low-to-high homophily: \method-2 achieves the best average rank (3.3) across all datasets (or homophily ratios $h$), followed by \method-1 (3.6).} %

\vspace{-0.25cm}
\paragraph{Additional model comparison} %
In Table~\ref{tab:5-real-rank}, we also report the \textit{best} results among the \textit{three} recently-proposed GEOM-GCN variants (\S~\ref{sec:related}), directly from the paper~\cite{Pei2020Geom-GCN}: %
other models (including ours) outperform this method significantly under heterophily.
We note that MLP is a competitive baseline under heterophily {(ranked 6.2)},
indicating that {many} existing models do not use the graph information effectively, or the latter is misleading in such cases. 
All models perform poorly on {\texttt{Squirrel} and \texttt{Actor}} likely due to their low-quality node features (small correlation with class labels). Also, \texttt{Squirrel} and \texttt{Chameleon} are dense, with many nodes sharing the same neighbors.  %

\vspace{-0.2cm}
\section{Conclusion}
\vspace{-0.2cm}

We have focused on characterizing the representation power of GNNs in challenging settings with heterophily or low homophily, which is understudied in the literature.
We have highlighted the current limitations of GNNs, 
presented designs that increase representation power under heterophily  %
and are theoretically justified with %
perturbation analysis and graph signal processing, and 
introduced the \method model that adapts to both heterophily and homophily by effectively synthetizing these designs.
We analyzed various challenging datasets, going beyond the often-used benchmark datasets (Cora, Pubmed, Citeseer), and leave as future work extending to a larger-scale experimental testbed.

\section*{Broader Impact} %

Homophily and heterophily are not intrinsically ethical or unethical---they are both phenomena existing in the nature, resulting in the popular proverbs %
``birds of a feather flock together'' and ``opposites attract''. 
However, many popular GNN models implicitly assume homophily; %
as a result, if they are applied to networks that do not satisfy the assumption, the results may be biased, unfair, or erroneous.
In some applications, the homophily assumption may have ethical implications. For example, a GNN model that intrinsically assumes homophily may contribute to the so-called ``filter bubble'' %
phenomenon in a recommendation system (reinforcing existing beliefs/views, and downplaying the opposite ones), 
or make minority groups less visible in social networks. 
In other cases, a reliance on homophily may hinder scientific progress. %
Among other domains, this is critical for applying GNN models to molecular and protein structures, where the connected nodes often belong to different classes, and thus successful methods will need to model heterophily successfully.%

Our work has the potential to rectify some of these potential negative consequences of existing GNN work.  While our methodology does not change the amount of homophily in a network, moving beyond a reliance on homophily can be a key to improve the fairness, diversity and performance in applications using GNNs.  
We hope that this paper will raise more awareness and discussions regarding the homophily limitations of existing GNN models, and help researchers design models which have the power of learning in both homophily and heterophily settings.

\begin{ack}
We thank the reviewers for their constructive feedback. 
This material is based upon work supported by the National Science Foundation under CAREER Grant No.~IIS 1845491 and 1452425, Army Young Investigator Award No.~W911NF1810397, an Adobe Digital Experience research faculty award, an Amazon faculty award, a Google faculty award, and AWS Cloud Credits for Research. We gratefully acknowledge the support of NVIDIA Corporation with the donation of the Quadro P6000 GPU used for this research. 
Any opinions, findings, and conclusions or recommendations expressed in this material are those of the author(s) and do not necessarily reflect the views of the National Science Foundation or other funding parties.

\end{ack}

{
\small
\bibliographystyle{BIB/ACM-Reference-Format}
\bibliography{BIB/abbreviations,BIB/ACM-abbreviations,BIB/main,BIB/all}
}

\appendix
\newpage
\numberwithin{table}{section}

\newpage
\section{Nomenclature}
\label{app:dfn}

We summarize the main symbols used in this work and their definitions below:

\begin{table}[h!]
     \caption{Major symbols and definitions.} %
     \label{tab:dfn}
     \resizebox{\textwidth}{!}{
    \begin{tabular}{ lp{11cm}}
         \toprule
         \textbf{Symbols} &  \textbf{Definitions}\\
         \midrule
         $\graph = (\vertexSet, \edgeSet)$ & graph $\graph$ with nodeset $\vertexSet$, edgeset $\edgeSet$ \\
         $\matA$ & $n \times n$ adjacency matrix of $\graph$ \\
         $\matX$ & $n \times F$ node feature matrix of $\graph$ \\
         $\V{x}_v$ & $F$-dimensional feature vector for node $v$ \\
         $\matL$ & unnormalized graph Laplacian matrix \\ 
         \midrule
         $\setY$ & set of class labels\\
         $y_v$ & class label for node $v \in \vertexSet$ \\
         $\vecy$ & $n$-dimensional vector of class labels (for all the nodes)\\
         \midrule
         $\setT = \{(v_1,y_1), (v_2, y_2), ...\}$ & training data for semi-supervised node classification \\
         $N(v)$ & general type of neighbors of node $v$ in graph $\graph$ \\
         $\neighNoSelfLoop(v)$ & general type of neighbors of node $v$ in $\graph$  \textit{without self-loops} (i.e., excluding $v$)  \\
         $N_i(v),\neighNoSelfLoop_i(v)$ & $i$-hop/step neighbors of node $v$ in $\graph$ (at exactly distance $i$) maybe-with/without self-loops, resp.\\
         $\edgeSet_2$ & set of pairs of nodes $(u,v)$ with shortest distance between them being 2 \\
         $d, d_{\mathrm{max}}$ & node degree, and maximum node degree across all nodes $v \in \vertexSet$, resp. \\
         \midrule
         $h$ & edge homophily ratio \\
         $\matH$ & class compatibility matrix \\
        \midrule
         $\V{r}^{(k)}_v$ & node representations learned in GNN model at round / layer $k$ \\
         $K$ & the number of rounds in the neighborhood aggregation stage \\
         $\matW$ & learnable weight matrix for GNN model \\
         $\sigma$ & non-linear activation function \\
         $\Vert$ & vector concatenation operator \\
         \texttt{AGGR} & function that aggregates node feature representations within a neighborhood \\
         \texttt{COMBINE} & function that combines feature representations from different neighborhoods \\
         \bottomrule
    \end{tabular}
    }
    \end{table}

\section{Homophily and Heterophily: Compatibility Matrix}
\label{app:homophily}

As we mentioned in \S~\ref{sec:homophily}, the edge homophily ratio in Definition~\ref{dfn:homophily-ratio} gives an \textit{overall trend} for all the edges in the graph. The actual level of homophily may vary within different pairs of node classes, i.e., there is different tendency of connection between each pair of classes. For instance, in an online purchasing network~\cite{netprobe07} with three classes---fraudsters, accomplices, and honest users---, fraudsters connect with higher probability to accomplices and honest users. %
Moreover, within the same network, it is possible that some pairs of classes exhibit homophily, while others exhibit heterophily. 
In belief propagation~\cite{yedidia2003understanding-fc}, a message-passing algorithm used for inference on graphical models, the different levels of homophily or affinity between classes are captured via the \textit{class compatibility}, \textit{propagation} or \textit{coupling} matrix, which is typically pre-defined based on domain knowledge.
In this work, we define the empirical \textit{class compatibility matrix} $\matH$ as follows: %
\begin{definition}
The class compatibility matrix $\matH$ has entries $[\matH]_{i,j}$ that capture the fraction of outgoing edges from a node in class $i$ to a node in class $j$: 
\begin{equation*}
    [\matH]_{i,j} = \frac{|\{(u,v): (u,v) \in \edgeSet \wedge y_u = i \wedge y_v = j\}|}{|\{(u,v): (u,v) \in \edgeSet \wedge y_u = i\}|}
\end{equation*}
\end{definition}
By definition, the class compatibility matrix is a stochastic matrix, with each row summing up to 1.

\newpage
\section{Proofs and Discussions of Theorems}

\subsection{Detailed Analysis of Theorem 1}
\label{app:proof-thm1}

\begin{proof}[for Theorem~\ref{thm:1}]
We first discuss the GCN layer formulated as $f(\matX; \matA, \matW) = (\matA+\matI)\matX\matW$. Given training set $\setT$, the goal of the training process is to optimize the weight matrix $\matW$ to minimize the loss function $\mathcal{L}([(\matA+\matI)\matX]_{\setT, :}\matW, [\V{Y}]_{\setT, :})$, where $[\V{Y}]_{\setT, :}$ is the one-hot encoding of class labels provided in the training set, and $[(\matA+\matI)\matX]_{\setT, :}\matW$ is the predicted probability distribution of class labels for each node $v$ in the training set $\setT$. 

Without loss of generality, we reorder $\setT$ accordingly such that the one-hot encoding of labels for nodes in training set $[\V{Y}]_{\setT, :}$ is in increasing order of the class label $y_v$: 
\begin{equation}
    \label{eq:app-proof-1-y}
    [\V{Y}]_{\setT, :} =
    \left[
    \begin{array}{ccccc}
    1 & 0 & 0 & \cdots & 0 \\ 
    \vdots & \vdots & \vdots & \ddots & \vdots \\
    1 & 0 & 0 & \cdots & 0 \\[6pt] \hdashline\noalign{\vskip 4pt}
    0 & 1 & 0 & \cdots & 0 \\ 
    \vdots & \vdots & \vdots & \ddots & \vdots \\
    0 & 1 & 0 & \cdots & 0 \\[6pt] \hdashline\noalign{\vskip 4pt}
    \vdots & \vdots & \vdots & \ddots & \vdots \\[6pt] \hdashline\noalign{\vskip 4pt}
    0 & 0 & 0 & \cdots & 1 \\ 
    \vdots & \vdots & \vdots & \ddots & \vdots \\
    0 & 0 & 0 & \cdots & 1 \\
    \end{array} 
    \right]_{|\vertexSet| \times |\setY|}
\end{equation}

Now we look into the term $[(\matA+\matI)\matX]_{\setT, :}$, which is the aggregated feature vectors within neighborhood $N_1$ for nodes in the training set. Since we assumed that all nodes in $\setT$ have degree $d$, proportion $h$ of their neighbors belong to the same class, while proportion $\tfrac{1-h}{|\setY|-1}$ of them belong to any other class uniformly, and one-hot representations of node features $\V{x}_v = \mathrm{onehot}(y_v)$ for each node $v$, we obtain:

\begin{equation}
    \label{eq:app-proof-1-aix}
    [(\matA+\matI)\matX]_{\setT, :} = \left[
    \begin{array}{ccccc}
        h d + 1 & \frac{1-h}{|\setY|-1} d & \frac{1-h}{|\setY|-1} d & \cdots & \frac{1-h}{|\setY|-1} d \\
        \vdots & \vdots & \vdots & \ddots & \vdots \\
        h d + 1 & \frac{1-h}{|\setY|-1} d & \frac{1-h}{|\setY|-1} d & \cdots & \frac{1-h}{|\setY|-1} d \\[6pt] \hdashline\noalign{\vskip 4pt}
        \frac{1-h}{|\setY|-1} d & h d + 1 &  \frac{1-h}{|\setY|-1} d & \cdots & \frac{1-h}{|\setY|-1} d \\
        \vdots & \vdots & \vdots & \ddots & \vdots \\
        \frac{1-h}{|\setY|-1} d & h d + 1 &  \frac{1-h}{|\setY|-1} d & \cdots & \frac{1-h}{|\setY|-1} d \\[6pt] \hdashline\noalign{\vskip 4pt}
        \vdots & \vdots & \vdots & \ddots & \vdots \\[6pt] \hdashline\noalign{\vskip 4pt}
        \frac{1-h}{|\setY|-1} d  &  \frac{1-h}{|\setY|-1} d  & \frac{1-h}{|\setY|-1} d & \cdots & h d + 1 \\
        \vdots & \vdots & \vdots & \ddots & \vdots \\
        \frac{1-h}{|\setY|-1} d  &  \frac{1-h}{|\setY|-1} d  & \frac{1-h}{|\setY|-1} d & \cdots & h d + 1 \\[6pt] 
    \end{array}
    \right]_{|\vertexSet| \times |\setY|}
\end{equation}

For $[\V{Y}]_{\setT, :}$ and $[(\matA+\matI)\matX]_{\setT, :}$ that we derived in Eq. \eqref{eq:app-proof-1-y} and \eqref{eq:app-proof-1-aix}, we can find an optimal weight matrix $\matW_*$ such that $[(\matA+\matI)\matX]_{\setT, :}\matW_* = [\V{Y}]_{\setT, :}$, making the loss  $\mathcal{L}([(\matA+\matI)\matX]_{\setT, :}\matW_*, [\V{Y}]_{\setT, :}) = 0$. We can use the following way to find $\matW_*$: First, sample one node from each class to form a smaller set $\mathcal{T}_S \subset \setT$, therefore we have: 
\begin{equation*}
    [\V{Y}]_{\mathcal{T}_S, :} =
    \left[
    \begin{array}{ccccc}
    1 & 0 & 0 & \cdots & 0 \\ 
    0 & 1 & 0 & \cdots & 0 \\ 
    \vdots & \vdots & \vdots & \ddots & \vdots \\
    0 & 0 & 0 & \cdots & 1 \\
    \end{array} 
    \right]_{|\setY| \times |\setY|} = \matI_{|\setY| \times |\setY|}
\end{equation*}
and
\begin{equation*}
    [(\matA+\matI)\matX]_{\mathcal{T}_S, :} = \left[
    \begin{array}{ccccc}
        h d + 1 & \frac{1-h}{|\setY|-1} d & \frac{1-h}{|\setY|-1} d & \cdots & \frac{1-h}{|\setY|-1} d \\
        \frac{1-h}{|\setY|-1} d & h d + 1 &  \frac{1-h}{|\setY|-1} d & \cdots & \frac{1-h}{|\setY|-1} d \\
        \vdots & \vdots & \vdots & \ddots & \vdots \\
        \frac{1-h}{|\setY|-1} d  & \frac{1-h}{|\setY|-1} d  & \frac{1-h}{|\setY|-1} d & \cdots & h d + 1 \\
    \end{array}
    \right]_{|\setY| \times |\setY|}
\end{equation*}

Note that $[(\matA+\matI)\matX]_{\mathcal{T}_S, :}$ is a circulant matrix, therefore its inverse exists. Using the Sherman-Morrison formula, we can find its inverse as:
{\small
\begin{align*}
    \left([(\matA+\matI)\matX]_{\mathcal{T}_S, :}\right)^{-1} 
     = & \frac{1}{(d+1)(|\setY| - 1 + (|\setY| h - 1)d)} \cdot \nonumber \\
     & 
    \resizebox{0.78\textwidth}{!}{$
    \left[ 
    \begin{array}{cccc}
        (|\setY| - 1) + (|\setY| - 2 + h)d & (h-1) d & \cdots & (h-1) d \\
        (h-1) d & (|\setY| - 1) + (|\setY| - 2 + h)d & \cdots & (h-1) d \\
        \vdots & \vdots & \ddots & \vdots \\
        (h-1) d & (h-1) d & \cdots & (|\setY| - 1) + (|\setY| - 2 + h)d
    \end{array} 
    \right]$}
\end{align*}
}%
Let $\matW_* = \left([(\matA+\matI)\matX]_{\mathcal{T}_S, :}\right)^{-1}$, and we have $[(\matA+\matI)\matX]_{\mathcal{T}_S, :} \matW_* = [\V{Y}]_{\mathcal{T}_S, :}=\matI_{|\setY| \times |\setY|}$. It is also easy to verify that $[(\matA+\matI)\matX]_{\setT, :}\matW_* = [\V{Y}]_{\setT, :}$. $\matW_* = \left([(\matA+\matI)\matX]_{\mathcal{T}_S, :}\right)^{-1}$ is the optimal weight matrix we can learn under $\setT$, since it satisfies $\mathcal{L}([(\matA+\matI)\matX]_{\setT, :}\matW_*, [\V{Y}]_{\setT, :}) = 0$.

Now consider an arbitrary training datapoint $(v, y_v) \in \setT$, and a perturbation added to the neighborhood $N(v)$ of node $v$, such that the number of nodes with a randomly selected class label $y_p \in \setY \neq y_v$ is $\delta_1$ less than expected in $N(v)$. We denote the perturbed graph adjacency matrix as $\matA_\Delta$. Without loss of generality, we assume node $v$ has $y_v = 1$, and the perturbed class is $y_p = 2$. In this case we have
\begin{equation*}
    [(\matA_\Delta + \matI)\matX]_{v, :} = \left[\begin{array}{ccccc}
        h d + 1 & \frac{1-h}{|\setY|-1} d - \delta_1 & \frac{1-h}{|\setY|-1} d & \cdots & \frac{1-h}{|\setY|-1} d \\
    \end{array}\right]
\end{equation*}
Applying the optimal weight matrix we learned on $\setT$ to the aggregated feature on the perturbed neighborhood $[(\matA_\Delta + \matI)\matX]_{v, :}$, we obtain $[(\matA_\Delta + \matI)\matX]_{v, :}\matW_*$ which equals to:
\begin{equation*}
    \resizebox{\textwidth}{!}{
    $\left[
    \begin{array}{ccccc}
        1 - \frac{(h-1) d \delta_1}{(d + 1)(|\setY| - 1 + (|\setY| h - 1)d)}
        & -\frac{((|\setY| - 1) + (|\setY| - 2 + h)d)\delta_1}{(d + 1)(|\setY| - 1 + (|\setY| h - 1)d)} 
        & - \frac{(h-1)d\delta_1}{(d + 1)(|\setY| - 1 + (|\setY| h - 1)d)}
        & \cdots &
        - \frac{(h-1)d\delta_1}{(d + 1)(|\setY| - 1 + (|\setY| h - 1)d)} \\
    \end{array}
    \right]$
    }
\end{equation*}
Notice that we always have $1 - \frac{(h-1) d \delta_1}{(d + 1)(|\setY| - 1 + (|\setY| h - 1)d)}$ > $- \frac{(h-1)d\delta_1}{(d + 1)(|\setY| - 1 + (|\setY| h - 1)d)}$, thus the GCN layer formulated as $(\matA+\matI)\matX\matW$ would misclassify only if the following inequality holds: 
\begin{equation*}
    1 - \frac{(h-1) d \delta_1}{(d + 1)(|\setY| - 1 + (|\setY| h - 1)d)} < -\frac{((|\setY| - 1) + (|\setY| - 2 + h)d)\delta_1}{(d + 1)(|\setY| - 1 + (|\setY| h - 1)d)}
\end{equation*}
Solving the above inequality for $\delta_1$, we get the amount of perturbation needed as
\begin{equation}
    \label{eq:app-proof-1-gcnloop-delta}
    \begin{cases}
        \delta_1 > \frac{-h |\setY| d-|\setY|+d+1}{|\setY|-1}, \text{when } 0 \leq h<\frac{-|\setY|+d+1}{|\setY| d} \\
        \delta_1 < \frac{-h |\setY| d-|\setY|+d+1}{|\setY|-1}, \text{when } h > \frac{-|\setY|+d+1}{|\setY| d} \\
    \end{cases}
\end{equation}
and the least absolute amount of perturbation needed is $|\delta_1| = |\frac{-h |\setY| d-|\setY|+d+1}{|\setY|-1}|$. \\

Now we move on to discuss the GCN layer formulated as $f(\matX; \matA, \matW) = \matA\matX\matW$ without self loops. Following similar derivations, we obtain the optimal weight matrix $\matW_*$ which makes $\mathcal{L}([\matA\matX]_{\setT, :}\matW_*, [\V{Y}]_{\setT, :}) = 0$ as:
\begin{align}
    \matW_* = \left([\matA\matX]_{\mathcal{T}_S, :}\right)^{-1} = & \frac{1}{(1 - h|\setY|) d} 
    \left[ 
    \begin{array}{cccc}
        - (|\setY| - 2 + h) & 1-h & \cdots & 1-h \\
        1-h & - (|\setY| - 2 + h) &  \cdots & 1-h \\
        \vdots & \vdots & \ddots & \vdots \\
        1 - h & 1 - h & \cdots & -(|\setY| - 2 + h) \\
    \end{array} 
    \right]
\end{align}

Again if for an arbitrary $(v, y_v) \in \setT$, a perturbation is added to the neighborhood $N(v)$ of the node $v$, such that the number of nodes with a randomly selected class label $y_p \in \setY \neq y_v$ is $\delta_2$ less than expected in $N(v)$, we have:
\begin{equation*}
    [\matA_\Delta\matX]_{v, :} = \left[\begin{array}{ccccc}
        h d & \frac{1-h}{|\setY|-1} d - \delta_2 & \frac{1-h}{|\setY|-1} d & \cdots & \frac{1-h}{|\setY|-1} d \\
    \end{array}\right]
\end{equation*}
Then applying the optimal weight matrix that we learned on $\setT$ to the aggregated feature on perturbed neighborhood $[\matA_\Delta\matX]_{v, :}$, we obtain $[\matA_\Delta\matX]_{v, :}\matW_*$ which equals to:
\begin{equation*}
    \left[
    \begin{array}{ccccc}
        1-\frac{ (1-h) \delta_2}{(1-h |\setY|) d}
        & \frac{(|\setY|-2+h)\delta_2}{(1-h |\setY|) d}
        & -\frac{ (1-h) \delta_2}{(1-h |\setY|) d}
        & \cdots 
        & -\frac{ (1-h) \delta_2}{(1-h |\setY|) d} \\
    \end{array}
    \right]
\end{equation*}
Thus, the GCN layer formulated as $\matA\matX\matW$ would misclassify when the following inequality holds: 
\begin{equation*}
    1-\frac{ (1-h) \delta_2}{(1-h |\setY|) d}
    < 
    \frac{(|\setY|-2+h)\delta_2}{(1-h |\setY|) d}
\end{equation*}
Or the amount of perturbation is:
\begin{equation}
    \label{eq:app-proof-1-gcn-delta}
    \begin{cases}
        \delta_2 > \frac{(1-h|\setY|)d}{|\setY|-1}, \text{when } 0 \leq h < \frac{1}{|\setY|} \\
        \delta_2 < \frac{(1-h|\setY|)d}{|\setY|-1}, \text{when } h > \frac{1}{|\setY|} \\
    \end{cases}
\end{equation}
As a result, the least absolute amount of perturbation needed is $|\delta_2| = |\frac{(1-h|\setY|)d}{|\setY|-1}|$. \\

By comparing the least absolute amount of perturbation needed for both formulations to misclassify ($|\delta_1| = |\frac{-h |\setY| d-|\setY|+d+1}{|\setY|-1}|$ derived in Eq. \eqref{eq:app-proof-1-gcnloop-delta} for the $(\matA+\matI)\matX\matW$ formulation; $|\delta_2| = |\frac{(1-h|\setY|)d}{|\setY|-1}|$ derived in Eq. \eqref{eq:app-proof-1-gcn-delta} for the $\matA\matX\matW$ formulation), 
we can see that $|\delta_1| = |\delta_2|$ if and only if $\delta_1 = -\delta_2$, which happens when $h = \frac{1-|\setY|+2d}{2 |\setY| d}$. When $h <  \frac{1-|\setY|+2d}{2 |\setY| d}$ (heterophily), we have $|\delta_1| < |\delta_2|$, which means the $(\matA+\matI)\matX\matW$ formulation is less robust to perturbation than the $\matA\matX\matW$ formulation. 
\hfill $\blacksquare$
\end{proof}

\begin{figure}[h]
    \centering
    \begin{subfigure}{0.3\textwidth}
        \centering
        \includegraphics[keepaspectratio, width=0.95\textwidth]{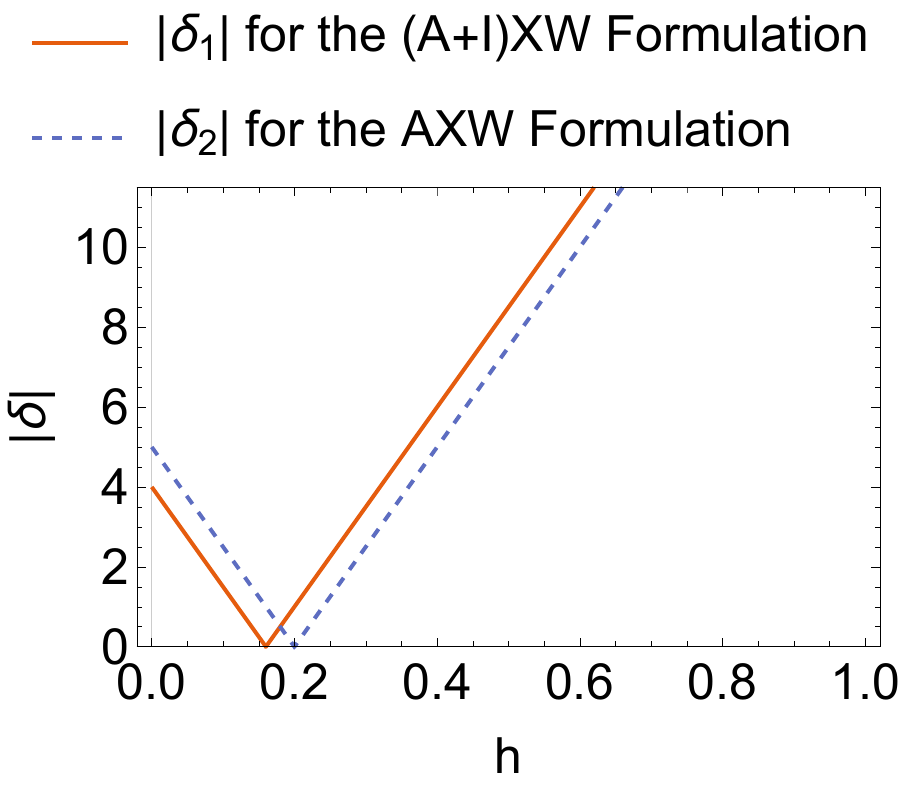}
        \caption{$|\delta|$ as a function of $h$ under $d=20, |\setY| = 5$.} 
        \label{fig:app-proof-1-discussion-h}
    \end{subfigure}
    \hfill
    \begin{subfigure}{0.3\textwidth}
        \centering
        \includegraphics[keepaspectratio, width=0.95\textwidth]{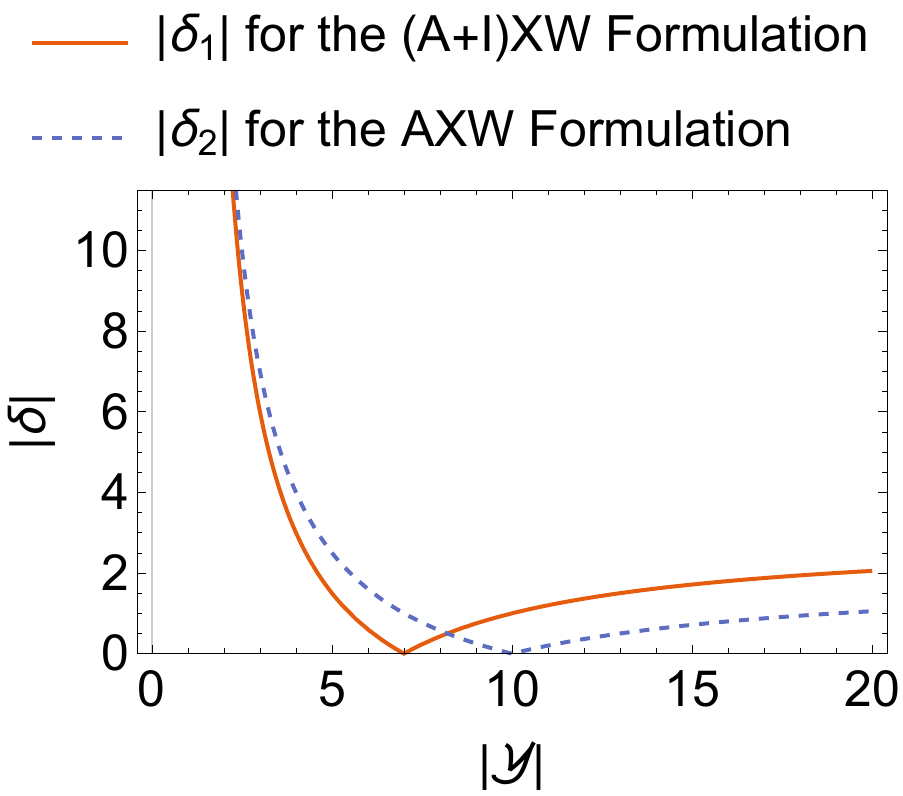}
        \caption{$|\delta|$ as a function of $|\setY|$ under $h=0.1, d=20$.} 
        \label{fig:app-proof-1-discussion-numclass}
    \end{subfigure}
    \hfill
    \begin{subfigure}{0.3\textwidth}
        \centering
        \includegraphics[keepaspectratio, width=0.95\textwidth]{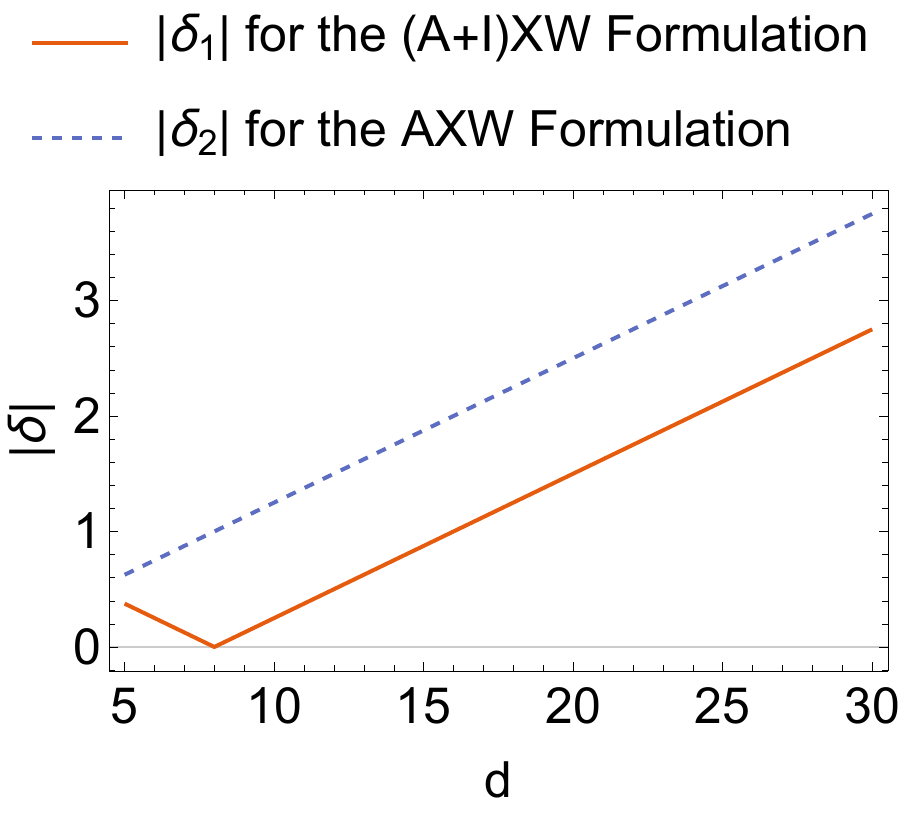}
        \caption{$|\delta|$ as a function of $d$ under $h=0.1, |\setY| = 5$.} 
        \label{fig:app-proof-1-discussion-d}
    \end{subfigure}
    \caption{Perturbation $|\delta|$ needed in order for GCN layers $(\matA+\matI)\matX\matW$ and $\matA\matX\matW$ to misclassify a node: Examples of perturbation $|\delta|$ as functions of $h$, $|\setY|$ and $d$, respectively.}
    \label{fig:app-proof-1-discussion}
\end{figure}

\paragraph{Discussions} From the above proof, we can see that the least absolute amount of perturbation $|\delta|$ needed for both GCN formulations is a function of the assumed homophily ratio $h$, the node degree $d$ for each node in the training set $\setT$, and the size of the class label set $|\setY|$. Fig. \ref{fig:app-proof-1-discussion} shows the plots of $|\delta_1|$ and $|\delta_2|$ as functions of $h$, $|\setY|$ and $d$: from Fig. \ref{fig:app-proof-1-discussion-h}, we can see that the least absolute amount of perturbations $|\delta|$ needed for both formulation first decreases as the assumed homophily level $h$ increases, until $\delta$ reaches 0, where the GCN layer predicts the same probability for all class labels; after that, $\delta$ decreases further below 0, and $|\delta|$ increases as $h$ increases; 
the $(\matA+\matI)\matX\matW$ formulation is less robust to perturbation than the $\matA\matX\matW$ formulation at low homophily level until $h=\frac{1-|\setY|+2d}{2 |\setY| d}$ as our proof shows, where $|\delta_1| = |\delta_2|$. Figure \ref{fig:app-proof-1-discussion-numclass} shows the changes of $|\delta|$ as a function of $|\setY|$ when fixed $h=0.1$ and $d=20$. For both formulations, $|\delta|$ first decrease rapidly as $|\setY|$ increases until $\delta$ reaches 0, after that $\delta$ increases slowly as $|\setY|$ increases; this reveals that both GCN formulations are more robust when $|\setY| << d$ under high homophily level, and in that case $\matA\matX\matW$ formulation is more robust than the $(\matA+\matI)\matX\matW$ formulation. Figure \ref{fig:app-proof-1-discussion-d} shows the changes of $|\delta|$ as a function of $d$ for fixed $h=0.1$ and $|\setY| = 5$: in this case the $\matA\matX\matW$ formulation is always more robust than the $(\matA+\matI)\matX\matW$ formulation, and for the $(\matA+\matI)\matX\matW$ formulation, $|\delta|$ follows again a ``V''-shape curve as $d$ changes. 

\subsection{Detailed Analysis of Theorem 2}
\label{app:proof-thm2}

\begin{proof}[for Theorem~\ref{thm:2}]
For all $v \in \vertexSet$, since its neighbors' class labels $\{y_u: u \in N(v)\}$ are conditionally independent given $y_v$, we can define a matrix $\V{P}_v$ for each node $v$ as $[\V{P}_v]_{i, j} = P(y_u = j | y_v = i), \forall i, j \in \setY, u \in N(v)$. Following the assumption that for all $v \in \vertexSet$, $P(y_u=y_v|y_v) = h$, $P(y_u=y|y_v) = \frac{1-h}{|\setY|-1}, \forall y \neq y_v$, we have
\begin{equation}
    \label{eq:app-proof-2-mat-p}
    \V{P}_v = \V{P} = \left[
    \begin{array}{cccc}
        h & \frac{1-h}{|\setY|-1} & \cdots & \frac{1-h}{|\setY|-1} \\
        \frac{1-h}{|\setY|-1}  & h & \cdots &  \frac{1-h}{|\setY|-1} \\
        \vdots & \vdots & \ddots & \vdots \\
        \frac{1-h}{|\setY|-1}  & \frac{1-h}{|\setY|-1} & \cdots &  h \\
    \end{array}
    \right], \; \forall v \in \vertexSet
\end{equation}
Now consider node $w \in N_2(v)$, we have: 
\begin{equation}
    P(y_w = k | y_v = i) = \sum_{j\in|\setY|} P(y_w = k | y_u = j) P(y_u = j | y_v = i) = \sum_{j\in|\setY|} [\V{P}]_{j, k}[\V{P}]_{i, j} = \V{P}^2 
\end{equation}

Therefore, to prove that the 2-hop neighborhood $N_2(v)$ for any node $v\in \vertexSet$ is homophily-dominant in expectation (i.e. $P(y_w = i | y_v = i) \geq P(y_w = j | y_v = i), \forall j \in \setY \neq i, w\in N_2(v)$), we need to show that the diagonal entries $[\V{P}^2]_{i,i}$ of $\V{P}^2$ are larger than the off-diagonal entries $[\V{P}^2]_{i,j}$. 

Denote $\rho = \frac{1-h}{|\setY|-1}$. From Eq. \eqref{eq:app-proof-2-mat-p}, we have
\begin{equation}
    [\V{P}^2]_{i,i} = h^2 + (|\setY| - 1)\rho^2
\end{equation}
and for $i \neq j$
\begin{equation}
    [\V{P}^2]_{i,j} = 2h\rho + (|\setY| - 2)\rho^2 
\end{equation}

Thus, $$[\V{P}^2]_{i,i} - [\V{P}^2]_{i,j} = h^2 - 2h\rho + \rho^2 = (h - \rho)^2 \ge 0$$
with equality if and only if $h = \rho$, namely $h = \frac{1}{|\setY|}$. Therefore, we proved that the 2-hop neighborhood $N_2(v)$ for any node $v\in \vertexSet$ will always be homophily-dominant in expectation. 
\hfill $\blacksquare$
\end{proof}

\subsection{Detailed Analysis of Theorem 3}
\label{app:proof-thm3}

\paragraph{Preliminaries} We define unnormalized Laplacian matrix of graph $\graph$ as $\matL = \matD - \matA$, where $\matA \in \{0, 1\}^{|\vertexSet| \times |\vertexSet|}$ is the adjacency matrix and $\matD$ is the diagonal matrix with $[\V{D}]_{i,i} = \sum_{j} [\matA]_{i,j}$. Without loss of generality, since the eigenvalues $\{\lambda_i\}$ of $\matL$ are real and nonnegative \cite{shuman2013emerging}, we assume the following order for the eigenvalues of $\matL$: $0 = \lambda_0 < \lambda_1 \leq \lambda_2 \leq \cdots \leq \lambda_{|\vertexSet|-1} = \lambda_{max}$. Furthermore, since $\matL$ is real and symmetric, there exists a set of orthonormal eigenvectors $\{\V{v}_i\}$ that form a complete basis of $\R^{|\vertexSet|}$. This means that for any graph signal $\V{s} \in \R^{|\vertexSet|}$, where $\V{s}_u$ is the value of the signal on node $u\in\vertexSet$, it can be decomposed to a weighted sum of $\{\V{v}_i\}$. Mathematically, $\V{s}$ is represented as $\V{s}=\sum_{i=0}^{|\vertexSet|-1}c_{s,i}\V{v}_i$, where $c_{s,i}=\V{s}^\T\V{v}_i$. We regard $c_{s,i}$ as the coefficient of $\V{s}$ at frequency component $i$ and regard the coefficients at all frequencies components $\{c_{s,i}\}$ as the spectrum of signal $\V{s}$ with respect to graph $\graph$.
In the above order of the eigenvalues, $\lambda_i$ which are closer to 0 would correspond to lower-frequency components, and $\lambda_i$ which are closer to $\lambda_{max}$ would correspond to higher-frequency components. 
Interested readers are referred to \cite{shuman2013emerging} for further details regarding signal processing on graphs. 

The \textbf{smoothness score of a signal} $\V{s}$ on graph $\graph$, which measures the amount of changes of signal $\V{s}$ along the edges of graph $\graph$, can be defined using $\matL$ as 
\begin{equation}
    \label{eq:app-smoothness}
    \V{s}^\T\matL\V{s} = \sum_{i, j} \matA_{ij} (\V{s}_i - \V{s}_j)^2 = \sum_{u \in \vertexSet} \sum_{v \in N(u)} (\V{s}_u - \V{s}_v)^2.
\end{equation}

Then, for two eigenvectors $\V{v}_i$ and $\V{v}_j$ corresponding to eigenvalues $\lambda_i \leq \lambda_j$ of $\matL$, we have: 
\begin{equation}
    \V{v}^\T_i \matL \V{v}_i = \lambda_i \leq \lambda_j = \V{v}^\T_j \matL \V{v}_j \nonumber
\end{equation}
which means that $\V{v}_i$ is \textit{more smooth} than $\V{v}_j$. This matches our expectations that a lower-frequency signal on $\graph$ should have smaller smoothness score. The \textbf{smoothness score for arbitrary graph signal} $\V{s} \in \R^{|\vertexSet|}$ can be represented by its coefficients of each frequency component as: 
\begin{equation}
    \label{eq:app-smoothness-spectrum}
    \V{s}^\T\matL\V{s} 
    = \left(\sum_i c_{s,i} \V{v}_i\right) \matL \left(\sum_i c_{s,i} \V{v}_i\right) 
    = \sum_{i=0}^{|\vertexSet| - 1} c_{s,i}^2 \lambda_i 
\end{equation}
with the above preliminaries, we can define the following concept:
\begin{definition}
\label{def:app-laplacian-spectrum}
Suppose $\V{s} = \sum_{i=0}^{|\vertexSet|-1} c_{s,i}\V{v}_i$ and $\V{t} = \sum_{i=0}^{|\vertexSet|-1} c_{t,i}\V{v}_i$ are two graph signals defined on $\graph$. 
In the spectrum of the unnormalized graph laplacian $\matL$, graph signal $\V{s}$ has higher energy on high-frequency components than $\V{t}$ if there exists integer $ 0 < M \leq |\vertexSet| - 1$ such that $\sum_{i=M}^{|\vertexSet|-1} c_{s,i}^2 > \sum_{i=M}^{|\vertexSet|-1} c_{t,i}^2$. 
\end{definition}

\vspace{0.1cm}
Based on these preliminary definitions, we can now proceed with the proof of the theorem:
\begin{proof}[for Theorem~\ref{thm:3}]
\label{proof:thm3}
We first prove that for graph signals $\V{s}, \V{t} \in \{0, 1\}^{|\vertexSet|}$, edge homophily ratio $h_s < h_t$ if and only if $\V{s}^\T\matL\V{s} > \V{t}^\T\matL\V{t}$. Following Dfn.~\ref{dfn:homophily-ratio}, the edge homophily ratio for signal $\V{s}$ (similarly for $\V{t}$) can be calculated as: 
\begin{equation}
\label{eq:homophily-signal}
    h_s = \frac{1}{2|\edgeSet|} \sum_{u \in \vertexSet} \left( 
        d_u - \sum_{v \in N(v)}(\V{s}_u - \V{s}_v)^2
    \right) = 
    \frac{1}{2|\edgeSet|} \sum_{u \in \vertexSet} d_u - \frac{1}{2|\edgeSet|} \sum_{u \in \vertexSet} \sum_{v \in N(v)}(\V{s}_u - \V{s}_v)^2 
\end{equation}
Plugging this in Eq. \eqref{eq:app-smoothness}, we obtain:
\begin{equation}
    h_s = \frac{1}{2|\edgeSet|} \sum_{u \in \vertexSet} d_u - \frac{1}{2|\edgeSet|} \V{s}^\T\matL\V{s} = 1 - \frac{1}{2|\edgeSet|} \V{s}^\T\matL\V{s} \nonumber
\end{equation}
where $|\edgeSet|$ is the number of edges in $\graph$. From the above equation, we have
\begin{equation}
    h_s < h_t \; \Leftrightarrow \;  
    1 - \frac{1}{2|\edgeSet|} \V{s}^\T\matL\V{s} < 1 - \frac{1}{2|\edgeSet|} \V{t}^\T\matL\V{t} \; \Leftrightarrow \; 
    \V{s}^\T\matL\V{s} > \V{t}^\T\matL\V{t} \nonumber
\end{equation}
i.e. edge homophily ratio $h_s < h_t$ if and only if $\V{s}^\T\matL\V{s} > \V{t}^\T\matL\V{t}$. 

Next we prove that if $\V{s}^\T\matL\V{s} > \V{t}^\T\matL\V{t}$, then following Dfn.\ref{def:app-laplacian-spectrum}, signal $\V{s}$ has higher energy on high-frequency components than $\V{t}$. We prove this by contradiction: suppose integer $ 0 < M \leq |\vertexSet| - 1$ does not exist such that $\sum_{i=M}^{|\vertexSet|-1} c_{si}^2 > \sum_{i=M}^{|\vertexSet|-1} c_{ti}^2$ when $\V{s}^\T\matL\V{s} > \V{t}^\T\matL\V{t}$, then all of the following inequalities must hold, as the eigenvalues of $\matL$ satisfy $0 = \lambda_0 < \lambda_1 \leq \lambda_2 \leq \cdots \leq \lambda_{|\vertexSet|-1} = \lambda_{max}$: 
\begin{align*}
    0 = \lambda_0 ( c^2_{s,0} + c^2_{s,1} + c^2_{s,2} + \cdots + c^2_{s,|\vertexSet|-1}) & = \lambda_0 ( c^2_{t,0} + c^2_{t,1} + c^2_{t,2} + \cdots + c^2_{t,|\vertexSet|-1}) = 0\\
    (\lambda_1 - \lambda_0) ( c^2_{s,1} + c^2_{s,2} + \cdots + c^2_{s,|\vertexSet|-1}) & \leq (\lambda_1 - \lambda_0) ( c^2_{t,1} + c^2_{t,2} + \cdots + c^2_{t,|\vertexSet|-1}) \\
    (\lambda_2 - \lambda_1) ( c^2_{s,2} + \cdots + c^2_{s,|\vertexSet|-1}) & \leq (\lambda_2 - \lambda_1) ( c^2_{t,2} + \cdots + c^2_{t,|\vertexSet|-1}) \\
    &\mathrel{\makebox[\widthof{=}]{\vdots}} \\
    (\lambda_{|\vertexSet|-1} - \lambda_{|\vertexSet|-2}) c^2_{s,|\vertexSet|-1} & \leq (\lambda_{|\vertexSet|-1} - \lambda_{|\vertexSet|-2}) c^2_{t,|\vertexSet|-1}
\end{align*}
Summing over both sides of all the above inequalities, we have 
\begin{equation*}
    \lambda_0 \cdot c^2_{s,0} + \lambda_1 \cdot c^2_{s,1} + \lambda_2 \cdot c^2_{s,2} + \cdots + \lambda_{|\vertexSet|-1} \cdot c^2_{s,|\vertexSet|-1} \leq 
    \lambda_0 \cdot c^2_{t,0} + \lambda_1 \cdot c^2_{t,1} + \lambda_2 \cdot c^2_{t,2} + \cdots + \lambda_{|\vertexSet|-1} \cdot c^2_{t,|\vertexSet|-1}
\end{equation*}
i.e., $\sum_{i=0}^{|\vertexSet| - 1} c_{si}^2 \lambda_i \leq \sum_{i=0}^{|\vertexSet| - 1} c_{ti}^2 \lambda_i $. 
However, from Eq. \eqref{eq:app-smoothness-spectrum}, we should have
\begin{equation}
    \V{s}^\T\matL\V{s} > \V{t}^\T\matL\V{t} \; \Leftrightarrow \;  
    \sum_{i=0}^{|\vertexSet| - 1} c_{si}^2 \lambda_i > \sum_{i=0}^{|\vertexSet| - 1} c_{ti}^2 \lambda_i \nonumber
\end{equation}
which contradicts with the previous resulting inequality. %
Therefore, the assumption should not hold, and there must exist an integer  $ 0 < M \leq |\vertexSet| - 1$ such that $\sum_{i=M}^{|\vertexSet|-1} c_{si}^2 > \sum_{i=M}^{|\vertexSet|-1} c_{ti}^2$ when $\V{s}^\T\matL\V{s} > \V{t}^\T\matL\V{t}$. \hfill $\blacksquare$
\end{proof}

\paragraph{Extension of Theorem \ref{thm:3} to one-hot encoding of class label vectors} Theorem \ref{thm:3} discusses only the graph signal $\V{s}, \V{t} \in \{0, 1\}^{|\vertexSet|}$ with only 1 channel (i.e., with only 1 value assigned to each node). It is possible to generalize the theorem to one-hot encoding $\V{Y}_s, \V{Y}_t \in \{0, 1\}^{|\vertexSet| \times |\setY|}$ as graph signal with $|\setY|$-channels by modifying Dfn. \ref{def:app-laplacian-spectrum} as follows: 

\begin{definition}
\label{def:app-laplacian-spectrum-alt}
Suppose $\left[\V{Y}_s\right]_{:,j} = \sum_{i=0}^{|\vertexSet|-1} c_{s,j,i}\V{v}_i$ and $\left[\V{Y}_t\right]_{:,j} = \sum_{i=0}^{|\vertexSet|-1} c_{t,j,i} \V{v}_i$ are one-hot encoding of class label vector $\vecy_s, \vecy_t$ defined as graph signals on $\graph$, where $c_{s,j,i} = [\V{Y}_{s}]_{:,j}^\T\V{v}_i$ is the coefficient of the $j$th-channel of $\V{Y}_s$ at frequency component $i$. In the spectrum of the unnormalized graph laplacian $\matL$, graph signal $\V{Y}_s$ has higher energy on high-frequency components than $\V{Y}_t$ if there exists integer $ 0 < M \leq |\vertexSet| - 1$ such that $\sum_{i=M}^{|\vertexSet|-1} \sum_{j=1}^{\phi} c_{s,j,i}^2 > \sum_{i=M}^{|\vertexSet|-1} \sum_{j=1}^{\phi} c_{t,j,i}^2$. 
\end{definition}

Under this definition, we can prove Theorem 3 %
for one-hot encoding of class label vectors $\V{Y}_s, \V{Y}_t$ as before, with the modification that in this case we have for signal $\V{Y}_s$ (similarly for $\V{Y}_t$): 
\begin{equation}
    h_s = \frac{1}{4|\edgeSet|} \sum_{u \in \vertexSet} \left( 
        2 d_u - \sum_{v \in N(v)}\sum_{j=1}^{\phi}([\V{Y}_s]_{u,j} - [\V{Y}_t]_{v,j})^2 \nonumber
    \right) 
\end{equation}
instead of Eq.~\eqref{eq:homophily-signal}. The rest of the proof is similar to Proof \ref{proof:thm3}.

\section{Our \method model: Details}
\label{app:algo}

In this section, we give the pipeline and pseudocode of \method, elaborate on its differences from existing GNN models, and present a detailed analysis of its computational complexity. 

\subsection{Pseudocode \& Pipeline} 
\label{app:pseudocode}

\begin{figure}[h]
    \centering
    \includegraphics[width=\textwidth, keepaspectratio, trim={0.4cm 0.4cm 0.8cm 0.4cm},clip]{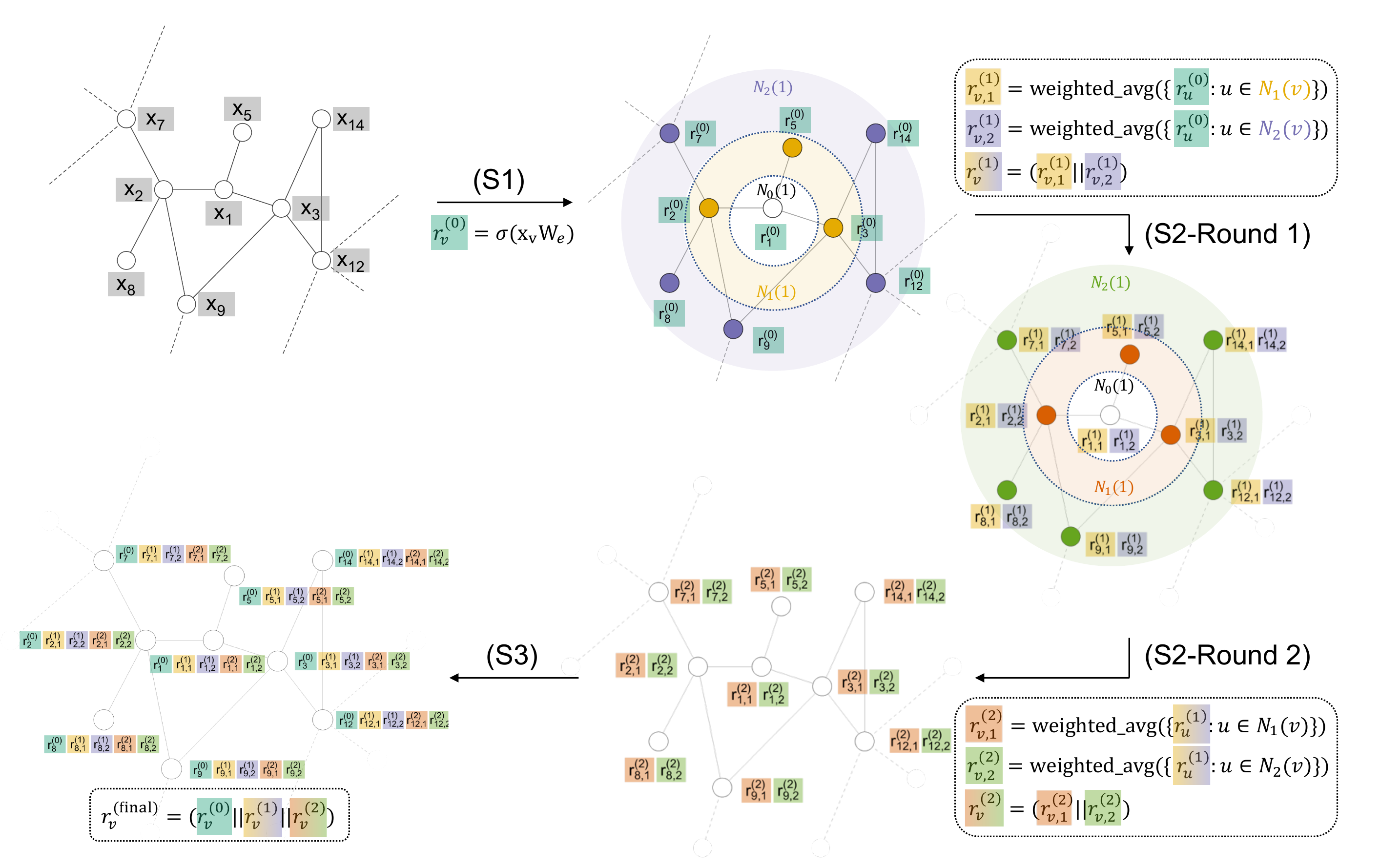}
    \caption{\method-2 pipeline. It consists of 3 stages: {\bf \stepone} feature embedding, {\bf \steptwo} neighborhood aggregation, and {\bf \stepthree} classification. 
    The \textit{feature embedding stage} {\bf \stepone} %
    uses a graph-agnostic dense layer to generate %
    the feature embedding {\small $\V{r}_v^{(0)}$} of each node $v$ based on its ego-feature $\V{x}_v$.
    In the \textit{neighborhood aggregation stage} {\bf \steptwo}, the generated embeddings are aggregated and repeatedly updated within the node's neighborhood; the 1-hop neighbors $N_1(v)$ and 2-hop neighbors $N_2(v)$ are aggregated separately and then concatenated, %
    following our design D2. 
    In the \textit{classification stage} {\bf \stepthree}, each node is classified based on its final embedding {\small $\V{r}_{v}^{(\text{final})}$}, which consists of its intermediate representations  concatenated as per design D3.
    }
    \label{fig:pipeline}
\end{figure}

In Fig.~\ref{fig:pipeline} we visualize \method, which we describe in \S~\ref{sec:method}. We also give its pseudocode in Algorithm~\ref{algo:method}. 

\begin{algorithm}[h!]
\caption{\method Framework for Node Classification under Homophily \& Heterophily}
\label{algo:method}
\SetNoFillComment
{\footnotesize 
\SetKwInput{Input}{Input}\SetKwInput{Output}{Output}
\SetKwInput{HyperParams}{Hyper-parameters}\SetKwInput{TrainParams}{Network Parameters}
\Input{%
Graph Adjacency Matrix $\matA \in \{0,1\}^{n\times n}$; 
Node Feature Matrix $\matX \in \mathbb{R}^{n \times F}$; 
Set of Labels $\setY$;
Labeled Nodes $\setT$
}
\HyperParams{
Dropout Rate; 
Non-linearity function $\sigma$; 
Number of Embedding Rounds $K$; 
Dimension of Feature Embedding $p$; %
}
\TrainParams{
$\V{W}_{e} \in \R^{F\times p}$; $\V{W}_{c} \in \mathbb{R}^{(2^{K+1}-1)p \times |\setY|}$
}
\Output{
Class label vector $\vecy$}
\Begin{

\tcc{All new variables defined below are initialized as all 0}

\vspace{0.15cm}

\tcc{{Stage S1: Feature Embedding}}
\For{$v \in \vertexSet$}{
    $\V{r}^{(0)}_{v} \leftarrow \sigma\left(\V{x}_{v}\V{W}_{e}\right)$ \tcc{Embeddings stored in matrix $\matR$}
}

\vspace{0.15cm}

\tcc{{Stage S2: Neighborhood Aggregation}}

\tcc{Calculate higher-order neighborhoods $\neighNoSelfLoop_1$ and $\neighNoSelfLoop_2$ \textbf{without self-loops} and their corresponding adjacency matrices $\bar{\matA}_1$ and $\bar{\matA}_2$}
$\matA_0 \leftarrow \matI_{n}$ \hfill \tcc{$\matI_n$ is the $n\times n$ identity matrix}
$\bar{\matA}_1 \leftarrow \mathbb{I}\left[\matA - \matI_n > 0  \right]$ \hfill  \tcc{$\mathbb{I}$ is a element-wise indicator function for matrix}
$\bar{\matA}_2 \leftarrow \mathbb{I}\left[\matA^2-\matA - \matI_n > 0 \right]$\;

\For{$i \leftarrow 1$ \KwTo $2$}{
    \For{$v \in \vertexSet$}{
            $d_{v,i} \leftarrow \sum_k {\bar{a}}_{vk, i}$ \hfill \tcc{degree of node $v$ at neighborhood $\neighNoSelfLoop_i$}
        }
        $\bar{{\matD}}_{i} \leftarrow \mathrm{diag}\{d_{v,i}: v \in \vertexSet \} $\;
        $\bar{{\matA}}_{i} \leftarrow \bar{\matD}_{i}^{-\frac{1}{2}} \bar{\matA}_{i} \bar{\matD}_{i}^{-\frac{1}{2}}$ \hfill \tcc{symmetric degree-normalization of matrices $\bar{\matA}_i$}
    }

\For{$k \leftarrow 1$ \KwTo $K$}{
    $\V{R}_1^{(k)} \leftarrow \bar{{\matA}}_{1} \V{R}^{(k-1)}$   \hfill   \tcc{Designs D1 + D2}
    $\V{R}_2^{(k)} \leftarrow \bar{{\matA}}_{2} \V{R}^{(k-1)}$\;
    \tcc{$\Vert$ is the vector concatenation operator}
        $\V{R}^{(k)} \leftarrow \left( \V{R}_1^{(k)} \Vert \V{R}_2^{(k)} \right) $ %
  }
$\V{R}^{(\text{final})} \leftarrow \left( \V{R}^{(0)} \Vert \V{R}^{(1)} \Vert \ldots \Vert \V{R}^{(K)} \right) $ \hfill \tcc{Design D3}

\vspace{0.15cm}

\tcc{Stage S3: Classification} 
$\V{R}^{(\text{final})} \leftarrow \mathrm{dropout}(\V{R}^{(\text{final})})$ \hfill \tcc{default dropout rate: 0.5}
\For{$v \in \vertexSet$}{
    $\V{p}_{v} \leftarrow \mathrm{softmax}(\V{r}_v^{(\text{final})} \V{W}_{c})$\;
    $\vecy_v \leftarrow \arg \max(\V{p}_{v})$ \hfill \tcc{class label}
    }
  }
  
}
\end{algorithm}

\subsection{Detailed Comparison of \method to existing GNN models} %
\label{app:qualitative-comp}

In \S~\ref{sec:related}, we discussed several high-level differences between \method and the various GNN models that we consider in this work, including the inclusion or not of designs D1-D3. Here we give some additional conceptual and mechanism differences.

As we have mentioned, \method differs from GCN~\cite{kipf2016semi} in a number of ways: 
(1) In each round of propagation/aggregation, GCN ``mixes'' the ego- and neighbor-representations by repeatedly averaging them to obtain the new node representations, while \method keeps them distinct via concatenation; 
(2) GCN considers only the 1-hop neighbors (including the ego / self-loops), while \method considers higher-order neighborhoods ($\neighNoSelfLoop_1$ and $\neighNoSelfLoop_2$); 
(3) GCN applies non-linear embedding transformations per round (e.g., RELU), while \method perform feature embedding for the ego in the first layer and drops all other non-linearities in the aggregation stage; and 
(4) GCN does not use the jumping knowledge framework (unlike \method), and makes the node classification predictions based on the last-round representations.

Unlike GAT, \method does not use any attention mechanism. Creating attention mechanisms that can generalize well to heterophily is an interesting future direction. 
Moreover, GCN-Cheby uses entirely different mechanisms than the other GNN models that we consider (i.e., Chebysev polynomials), though it has some conceptual similarities to \method in terms of the higher-order neighborhoods that it models.

GraphSAGE differs from \method in the same ways that are described in (2)-(4) above. In addition to leveraging only the 1-hop neighborhood, GraphSAGE also samples a fixed number of neighbors per round, while \method uses the full neighborhood. With respect to ego- and neighbor-representations, GraphSAGE concatenates them (as we do) but subsequently applies non-linear embedding transformations to them jointly (while we simplify all non-linear transformations). Our empirical analysis has revealed that such transformations lead to a decrease in performance in heterophily settings (see paragraph below on ``Non-linear embedding transformations...'').

Finally, MixHop differs from \method in the same ways that are described in (1) and (3)-(4) above. It explicitly considers higher-order neighborhoods up to $N_2$, though \cite{MixHop} defines the 2-hop neighborhoods as that including neighbors \textit{up to} 2-hop away neighbors. In our framework, we define the $i$-hop neighborhood as the set of neighbors with minimum distance exactly $i$ from the ego (\S~\ref{sec:preliminaries}). Finally, the output layer of MixHop uses a tailored, column-wise attention layer, which prioritizes specific features, before the softmax layer. In contrast, before the classification layer, \method uses concatenation-based jumping knowledge in order to represent the high-frequency components that are critical in heterophily.

\paragraph{Non-linear embedding transformations per round in \method?}  %
GCN~\cite{kipf2016semi}, GraphSAGE~\cite{hamilton2017inductive} and other GNN models embed the intermediate representations per round of feature propagation and aggregation. 
However, as we show in the ablation study in App.~\ref{app:synthetic-results} (Table~\ref{tab:design-ablations}, last row ``Non-linear''), introducing non-linear transformations per round of the neighborhood aggregation stage \steptwo of \method-2 (i.e., with $K=2$) as follows leads to \textit{worse} performance than the framework design that we introduce in Eq.~\eqref{eq:h2gcn-neighbor-combine} of \S~\ref{sec:method}:
{\small
\begin{equation}
    \V{r}^{(k)}_v =  \texttt{COMBINE}\left( \sigma \left(\matW \left[\V{r}^{(k-1)}_v, \; \texttt{AGGR}\{\V{r}^{(k-1)}_u: u \in N_1(v)\}, \texttt{AGGR}\{\V{r}^{(k-1)}_u: u \in N_2(v)\}\right]\right)\right), %
\end{equation}
}
where $\sigma$ is RELU and $\matW$ is a learnable matrix. Our design in Eq.~\ref{eq:h2gcn-neighbor-combine} aggregates different neighborhoods in a similar way to SGC~\cite{wu2019simplifying}, which has shown that removing non-linearities does not negatively impact performance in homophily settings. We actually find that removing non-linearities even \textit{improves} the performance under heterophily.

\subsection{\method: Time Complexity in Detail 
}
\label{sec:app-complexity-detail}

\paragraph{Preliminaries} %
The worst case time complexity for calculating $\matA \cdot \matB$ when both $\matA$ and $\matB$ are sparse matrices is $\mathrm{O}(\text{nnz}(\matA)\cdot c_\matB)$, where $\text{nnz}(\matA)$ is the number of non-zero elements in matrix $\matA$, and $c_\matB = \max(\sum_j \mathbb{I}[b_{ij} > 0])$ is the maximum number of non-zero elements in any row of matrix $\matB$. The time complexity for calculating $\matA \cdot \matX$, when $\matX$ is a dense matrix with $F$ columns, is $\mathrm{O}(\text{nnz}(\matA)F)$. 

\paragraph{Time complexity of \method} 
We analyze the time complexity of \method by stage (except the classification stage).

The feature embedding stage \stepone takes $\mathrm{O}(\text{nnz}(\matX)p)$ to calculate $\sigma(\matX \matW_e)$  %
where $\matW_e\in\mathbb{R}^{F\times p}$ is a  learnable dense weight matrix, and $\matX \in \mathbb{R}^{n \times F}$ is the node feature matrix.

In the neighborhood aggregation stage \steptwo, we perform the following computations:
\begin{itemize}
    \item \textit{Calculation of higher-order neighborhoods.} Given that $\matA$ is sparse, we can obtain the 2-hop neighborhood by calculating $\matA^2$ in $\mathrm{O}\left(|\edgeSet|d_{\mathrm{max}}\right)$, where $|\edgeSet|$ is the number of edges in $\graph$ (equal to the number of non-zeroes in $\matA$), and  $d_{\mathrm{max}}$ is the maximum degree across all nodes $v \in \vertexSet$ (which is equal to the maximum number of non-zeroes in any row of $\matA$). 
    \item \textit{Feature Aggregation.} We begin with a $p$-dimensional embedding for each node after feature embedding. 
    In round $k$, since we are using the neighborhoods $\bar{N}_1$ and $\bar{N}_2$, we have an embedding $\V{R}^{(k-1)} \in \mathbb{R}^{n \times 2^{(k-1)} p}$ as input. %
    We aggregate embedding vectors within neighborhood by $\V{R}^{(k)} = \left( \bar{{\matA}}_{1} \V{R}^{(k-1)} \Vert \bar{{\matA}}_{2} \V{R}^{(k-1)} \right)$, in which $\bar{{\matA}}_{i}$ corresponds to the adjacency matrix of neighborhood $\bar{N}_i$. The two sparse matrix-matrix multiplications in the concatenation take
    $\mathrm{O}\left(|\edgeSet|2^{(k-1)} p + |\edgeSet_2|2^{(k-1)} p \right)$, where $|\edgeSet_2| = \frac{1}{2}\sum_{v\in \vertexSet} |\neighNoSelfLoop_2(v)|$. 
    Over $K$ rounds of embedding, the complexity becomes $\mathrm{O}\left(2^K(|\edgeSet| + |\edgeSet_2|) p\right)$. 
\end{itemize}

Adding all the big-O terms above, we have the overall time complexity for stages \stepone and \steptwo of \method as:
$$ \mathrm{O}\left(\text{nnz}(\matX)\,p + |\edgeSet|d_{\mathrm{max}} + 2^K (|\edgeSet| + |\edgeSet_2|) p\right),$$
where $K$ is usually a small number (e.g., 2). For small values of $K$, the complexity becomes $ \mathrm{O}\left(|\edgeSet|d_{\mathrm{max}} + (\text{nnz}(\matX) + |\edgeSet| + |\edgeSet_2|) p\right)$.

\section{Additional Related Work}
\label{app:related}

In \S~\ref{sec:related}, we discuss relevant work on GNNs. Here we briefly mention other approaches for node classification. %

Collective classification in statistical relational learning focuses on the problem of node classification by leveraging the correlations between the node labels and their attributes~\cite{sen2008collective}. Since exact inference is NP-hard, approximate inference algorithms (e.g., iterative classification~\cite{NevilleJ00-ICA,QingGetoor03-ICA}, loopy belief propagation) are used to solve the problem. 
Belief propagation (BP)~\cite{yedidia2003understanding-fc} is a classic message-passing algorithm %
for graph-based semi-supervised learning, which can be used for graphs exhibiting homophily or heterophily~\cite{KoutraKKCPF11} and has fast linearized versions~\cite{GatterbauerGKF15,eswaran2017zoobp}.  Different from the setup where GNNs are employed, BP does \textit{not} {by itself} leverage node features, and {usually} assumes a pre-defined class compatibility or edge potential matrix (\S~\ref{sec:preliminaries}).
{We note, however, that \citet{gatterbauer2014semi} proposed estimating the class compatibility matrix instead of using a pre-defined one in the BP formulation.} 
{Moreover, the recent CPGNN model~\cite{zhu2020graph} integrates the compatibility matrix as a set of learnable parameters into GNN, which it initializes with an estimated class compatibility matrix.}
Another classic approach for %
collective classification or graph-based semi-supervised learning is label propagation, which iteratively propagates the (up-to-date) label information of each node to its neighbors in order to minimize the overall smoothness penalty of label assignments in the graph. 
Standard label propagation approaches inherently assume homophily by penalizing different label assignments among immediate neighborhoods, but more recent works have also looked into formulations which can better address heterophily: 
{Before applying label propagation, \citet{peel2017graph} transforms the original graph into either a similarity graph by measuring similarity between node neighborhoods or 
a new graph connecting nodes that are two hops away;  
\citet{chin2019decoupled} decouple graph smoothing where the notion of ``identity'' and ``preference'' for each node are considered separately.  
However, like BP, these approaches do not by themselves utilize node features.
}

\vspace{0.3cm}
\section{Experimental Setup \& Hyperparameter Tuning}
\label{app:tuning}

\subsection{Setup} 

\paragraph{\method Implementation} We use $K = 1$ for \method-1 and $K=2$ for \method-2. 
For loss function, we calculate the cross entropy between the predicted %
and the ground-truth labels for nodes within the training set, and add $L_2$ regularization of network parameters $\V{W}_{e}$ and $\V{W}_{c}$. (cf.\ Alg.~\ref{algo:method})

\paragraph{Baseline Implementations} For all baselines besides MLP, we used the official implementation released by the authors on GitHub. 
\begin{itemize}
    \item \textbf{GCN \& GCN-Cheby}~\cite{kipf2016semi}: \url{https://github.com/tkipf/gcn}
    \item \textbf{GraphSAGE}~\cite{hamilton2017inductive}: \url{https://github.com/williamleif/graphsage-simple} (PyTorch implementation)
    \item \textbf{MixHop}~\cite{MixHop}: \url{https://github.com/samihaija/mixhop}
    \item \textbf{GAT}~\cite{velickovic2018graph}: \url{https://github.com/PetarV-/GAT}. (For large datasets, we make use of the sparse version provided by the author.)
\end{itemize}

For MLP, we used our own implementation of MLP with 1-hidden layer, which is equivalent to the case of $K=0$ in Algorithm 1. We use the same loss function as \method for training MLP. 

\paragraph{Hardware Specifications} We run experiments on synthetic benchmarks with an Amazon EC2 instance with instance size as \texttt{p3.2xlarge}, which features an 8-core CPU, 61 GB Memory, and a Tesla V100 GPU with 16 GB GPU Memory. 
For experiments on real benchmarks, we use a workstation with a 12-core AMD Ryzen 9 3900X CPU, 64GB RAM, and a Quadro P6000 GPU with 24 GB GPU Memory.

\subsection{Tuning the GNN Models}

To avoid bias, we tuned the hyperparameters of each method (\method and baseline models) on each benchmark. 
Below we list the hyperparameters tested on each benchmark per model. 
As the hyperparameters defined by each baseline model differ significantly, we list the combinations of non-default command line arguments we tested, without explaining them in detail. %
We refer the interested reader to the corresponding original implementations for further details on the arguments, including their definitions. 

\paragraph{Synthetic Benchmark Tuning} For each synthetic benchmark, we report the results for different heterophily levels under the same set of hyperparameters for each method, so that we can compare how the same hyperparameters perform across the full spectrum of low-to-high homophily. 
We report the best performance, for the set of hyperparameters which performs the best on the validation set on the majority of the heterophily levels for each method. 

For \texttt{syn-cora}, we test the following command-line arguments for each baseline method: 
\begin{itemize}
    \item \textbf{\method-1 \& \method-2}: 
    \begin{itemize}
        \item Dimension of Feature Embedding $p$: 64
        \item Non-linearity Function $\sigma$: ReLU
        \item Dropout Rate: $a\in \{0, 0.5\}$
    \end{itemize}
    We report the best performance, for $a=0$. 
    \item \textbf{GCN}~\cite{kipf2016semi}: 
    \begin{itemize}
        \item \texttt{\lstinline{hidden1}}: $a\in \{16, 32, 64\}$
        \item \texttt{\lstinline{early_stopping}}: $b\in\{40, 100, 200\}$
        \item \texttt{\lstinline{epochs}}: 2000
    \end{itemize}
    We report the best performance, for $a=32, b=40$.
    \item \textbf{GCN-Cheby}~\cite{kipf2016semi}: 
    \begin{itemize}
        \item Set 1:
        \begin{itemize}
        \item \texttt{\lstinline{hidden1}}: $a\in\{16, 32, 64\}$
        \item \texttt{\lstinline{dropout}}: 0.6
        \item \texttt{\lstinline{weight_decay}}: $b\in\{\texttt{1e-5, 5e-4}\}$
        \item \texttt{\lstinline{max_degree}}: 2
        \item \texttt{\lstinline{early_stopping}}: 40
        \end{itemize}
        \item Set 2:
        \begin{itemize}
        \item \texttt{\lstinline{hidden1}}: $a$ $\in \{16, 32, 64\}$
        \item \texttt{\lstinline{dropout}}: 0.5
        \item \texttt{\lstinline{weight_decay}}: \texttt{5e-4}
        \item \texttt{\lstinline{max_degree}}: 3
        \item \texttt{\lstinline{early_stopping}}: 40
    \end{itemize}
    \end{itemize}
    We report the best performance, for Set 1 with $a=64, b=\texttt{5e-4}$.
    \item \textbf{GraphSAGE}~\cite{hamilton2017inductive}: 
    \begin{itemize}
        \item \texttt{\lstinline{hid_units}}: $a\in \{64, 128\}$
        \item \texttt{\lstinline{lr}}: $b\in \{0.1, 0.7\}$
        \item \texttt{\lstinline{epochs}}: 500
    \end{itemize}
    We report the performance with $a=64, b=0.7$.
    
    \item \textbf{MixHop}~\cite{MixHop}: 
    \begin{itemize}
        \item \texttt{\lstinline{hidden_dims_csv}}: $a \in \{64, 192\}$
        \item \texttt{\lstinline{adj_pows}}: 0, 1, 2
    \end{itemize}
    We report the performance with $a=192$.
    \item \textbf{GAT}~\cite{velickovic2018graph}: 
    \begin{itemize}
        \item \texttt{\lstinline{hid_units}}: $a \in \{8, 16, 32, 64\}$
        \item \texttt{\lstinline{n_heads}}: $b\in\{1, 4, 8\}$ 
    \end{itemize}
    We report the performance with $a=8, b=8$.
    \item \textbf{MLP}
    \begin{itemize}
        \item Dimension of Feature Embedding $p$: 64
        \item Non-linearity Function $\sigma$: ReLU
        \item Dropout Rate: 0.5
    \end{itemize}
\end{itemize}

For \texttt{syn-products}, we test the following command-line arguments for each baseline method: 
\begin{itemize}
    \item \textbf{\method-1 \& \method-2}: 
    \begin{itemize}
        \item Dimension of Feature Embedding $p$: 64
        \item Non-linearity Function $\sigma$: ReLU
        \item Dropout Rate: $a \in \{0, 0.5\}$
    \end{itemize}
    We report the best performance, for $a=0.5$. 
    \item \textbf{GCN}~\cite{kipf2016semi}: 
    \begin{itemize}
        \item \texttt{\lstinline{hidden1}}: 64
        \item \texttt{\lstinline{early_stopping}}: $a \in \{40, 100, 200\}$
        \item \texttt{\lstinline{epochs}}: 2000
    \end{itemize}
    In addition, we disabled the default feature normalization in the official implementation, as the feature vectors in this benchmark have already been normalized, and we found the default normalization method hurts the performance significantly. We report the best performance, for $a=40$.
    \item \textbf{GCN-Cheby}~\cite{kipf2016semi}: 
    \begin{itemize}
        \item \texttt{\lstinline{hidden1}}: 64
        \item \texttt{\lstinline{max_degree}}: 2
        \item \texttt{\lstinline{early_stopping}}: 40
        \item \texttt{\lstinline{epochs}}: 2000
    \end{itemize}
    We also disabled the default feature normalization in the official implementation for this baseline. %
    \item \textbf{GraphSAGE}~\cite{hamilton2017inductive}: 
    \begin{itemize}
        \item \texttt{\lstinline{hid_units}}: $a \in \{64, 128\}$
        \item \texttt{\lstinline{lr}}: $b \in \{0.1, 0.7\}$
        \item \texttt{\lstinline{epochs}}: 500
    \end{itemize}
    We report the performance with $a=128, b=0.1$.
    \item \textbf{MixHop}~\cite{MixHop}: 
    \begin{itemize}
        \item \texttt{\lstinline{hidden_dims_csv}}: $a \in \{64, 192\}$
        \item \texttt{\lstinline{adj_pows}}: 0, 1, 2
    \end{itemize}
    We report the performance with $a=192$.
    \item \textbf{GAT}~\cite{velickovic2018graph}: 
    \begin{itemize}
        \item \texttt{\lstinline{hid_units}}: $8$
    \end{itemize}
    We also disabled the default feature normalization in the official implementation for this baseline.
    \item \textbf{MLP}
    \begin{itemize}
        \item Dimension of Feature Embedding $p$: 64
        \item Non-linearity Function $\sigma$: ReLU
        \item Dropout Rate: 0.5
    \end{itemize}
\end{itemize}

\paragraph{Real Benchmark (except {Cora-Full}) Tuning} For each real benchmark in Table~\ref{tab:5-real-results} (except \textbf{Cora-Full}),  %
we perform hyperparameter tuning (see values below) and report the best performance of each method %
on the validation set. So, for each method, its performance on different benchmarks can be reported from different hyperparameters. We test the following command-line arguments for each baseline method: 
\begin{itemize}
    \item \textbf{\method-1 \& \method-2}: %
    \begin{itemize}
        \item Dimension of Feature Embedding $p$: 64
        \item Non-linearity Function $\sigma$: \{\texttt{ReLU, None}\}
        \item Dropout Rate: $\{0, 0.5\}$
        \item L2 Regularization Weight: \{\texttt{1e-5, 5e-4}\}
    \end{itemize}
    \item \textbf{GCN}~\cite{kipf2016semi}: %
    \begin{itemize}
        \item \texttt{\lstinline{hidden1}}: 64
        \item \texttt{\lstinline{early_stopping}}: $\{40, 100, 200\}$
        \item \texttt{\lstinline{epochs}}: 2000
    \end{itemize}
    \item \textbf{GCN-Cheby}~\cite{kipf2016semi}: %
    \begin{itemize}
        \item \texttt{\lstinline{hidden1}}: 64
        \item \texttt{\lstinline{weight_decay}}: $\{\texttt{1e-5, 5e-4}\}$
        \item \texttt{\lstinline{max_degree}}: 2
        \item \texttt{\lstinline{early_stopping}}: $\{40, 100, 200\}$
        \item \texttt{\lstinline{epochs}}: 2000
    \end{itemize}
    \item \textbf{GraphSAGE}~\cite{hamilton2017inductive}: %
    \begin{itemize}
        \item \texttt{\lstinline{hid_units}}: 64
        \item \texttt{\lstinline{lr}}: $\{0.1, 0.7\}$
        \item \texttt{\lstinline{epochs}}: 500
    \end{itemize}
    \item \textbf{MixHop}~\cite{MixHop}: %
    \begin{itemize}
        \item \texttt{\lstinline{hidden_dims_csv}}: $\{64, 192\}$
        \item \texttt{\lstinline{adj_pows}}: 0, 1, 2
    \end{itemize}
    \item \textbf{GAT}~\cite{velickovic2018graph}: 
    \begin{itemize}
        \item \texttt{\lstinline{hid_units}}: $8$
    \end{itemize}
    \item \textbf{MLP}
    \begin{itemize}
        \item Dimension of Feature Embedding $p$: 64
        \item Non-linearity Function $\sigma$: $\{\texttt{ReLU, None}\}$
        \item Dropout Rate: $\{0, 0.5\}$
    \end{itemize}
\end{itemize}

For \textbf{GCN+JK}, \textbf{GCN-Cheby+JK} and \textbf{GraphSAGE+JK}, %
we enhanced the corresponding base model with jumping knowledge (JK) connections using JK-Concat~\cite{XuLTSKJ18-jkn} \emph{without} changing the number of layers or other hyperparameters for the base method.

\paragraph{Cora Full Benchmark Tuning}
The number of class labels in Cora-Full are many more compared to the other benchmarks (Table~\ref{tab:5-real-results}), which 
leads to a significant increase in the size of training parameters for each model. Therefore, we need to re-tune the hyperparameters, especially the regularization weights and learning rates, in order to get reasonable performance. We test the following command-line arguments for each baseline method: 
\begin{itemize}
    \item \textbf{\method-1 \& \method-2}: %
    \begin{itemize}
        \item Dimension of Feature Embedding $p$: 64
        \item Non-linearity Function $\sigma$: $\{\texttt{ReLU, None}\}$
        \item Dropout Rate: $\{0, 0.5\}$
        \item L2 Regularization Weight: $\{\texttt{1e-5, 1e-6}\}$
    \end{itemize}
    \item \textbf{GCN}~\cite{kipf2016semi}: %
    \begin{itemize}
        \item \texttt{\lstinline{hidden1}}: 64
        \item \texttt{\lstinline{early_stopping}}: $\{40, 100, 200\}$
        \item \texttt{\lstinline{weight_decay}:} $\{\texttt{5e-5, 1e-5, 1e-6}\}$
        \item \texttt{\lstinline{epochs}}: 2000
    \end{itemize}
    \item \textbf{GCN-Cheby}~\cite{kipf2016semi}: %
    \begin{itemize}
        \item \texttt{\lstinline{hidden1}}: 64
        \item \texttt{\lstinline{weight_decay}}: $\{\texttt{5e-5, 1e-5, 1e-6}\}$
        \item \texttt{\lstinline{max_degree}}: 2
        \item \texttt{\lstinline{early_stopping}}: $\{40, 100, 200\}$
        \item \texttt{\lstinline{epochs}}: 2000
    \end{itemize}
    \item \textbf{GraphSAGE}~\cite{hamilton2017inductive}: %
    \begin{itemize}
        \item \texttt{\lstinline{hid_units}}: 64
        \item \texttt{\lstinline{lr}}: 0.7
        \item \texttt{\lstinline{epochs}}: 2000
    \end{itemize}
    \item \textbf{MixHop}~\cite{MixHop}: %
    \begin{itemize}
        \item \texttt{\lstinline{adj_pows}}: 0, 1, 2
        \item \texttt{\lstinline{hidden_dims_csv}}: $\{64, 192\}$
        \item \texttt{\lstinline{l2reg}}: $\{\texttt{5e-4, 5e-5}\}$
    \end{itemize}
    \item \textbf{GAT}~\cite{velickovic2018graph}: 
    \begin{itemize}
        \item \texttt{\lstinline{hid_units}}: $8$
        \item \texttt{\lstinline{l2_coef}}: $\{\texttt{5e-4, 5e-5, 1e-5}\}$
    \end{itemize}
    \item \textbf{MLP}
    \begin{itemize}
        \item Dimension of Feature Embedding $p$: 64
        \item Non-linearity Function $\sigma$: $\{\texttt{ReLU, None}\}$
        \item Dropout Rate: $\{0, 0.5\}$
        \item L2 Regularization Weight: \texttt{1e-5}
        \item Learning Rate: 0.05
    \end{itemize}
\end{itemize}

For \textbf{GCN+JK}, \textbf{GCN-Cheby+JK} and \textbf{GraphSAGE+JK}, %
we enhanced the corresponding base model with jumping knowledge (JK) connections using JK-Concat~\cite{XuLTSKJ18-jkn} \emph{without} changing the number of layers or other hyperparameters for the base method.

\newpage
\section{Synthetic Datasets: Details}
\label{app:synthetic}

\subsection{Data Generation Process \& Setup}

\paragraph{Synthetic graph generation} We generate synthetic graphs with various heterophily levels by adopting an approach similar to \cite{MixHop, karimi2017visibility17}. In general, the synthetic graphs are generated by a modified preferential attachment process \cite{Barabasi:1999}: The number of class labels $|\setY|$ in the synthetic graph is prescribed. Then, starting from a small initial graph, new nodes are added into the graph one by one, until the number of nodes $|\vertexSet|$ has reached the preset level. The probability $p_{uv}$ for a newly added node $u$ in class $i$ to connect with an existing node $v$ in class $j$ is proportional to both the class compatibility $H_{ij}$ between class $i$ and $j$, and the degree $d_v$ of the existing node $v$. As a result, the degree distribution for the generated graphs follow a power law, and the heterophily can be controlled by class compatibility matrix $\matH$. 
Table \ref{tab:5-synthetic-data} shows an overview of these synthetic benchmarks, and more detailed statistics %
can be found in Table~\ref{tab:A-synthetic-stats}. 

\paragraph{Node features \& classes} Nodes are assigned randomly to each class during the graph generation. Then, in each synthetic graph, the feature vectors of nodes in each class are generated by sampling feature vectors of nodes from the corresponding class in a real benchmark (e.g., Cora~\cite{sen2008collective, yang2016revisiting}  or \texttt{ogbn-products}~\cite{hu2020ogb}): 
We first establish a class mapping $\psi: \setY_s \rightarrow \setY_b$ between classes in the synthetic graph $\setY_s$ to classes in an existing benchmark $\setY_b$. The only requirement is that the class size in the existing benchmark is larger than that of the synthetic graph so that an injection between nodes from both classes can be established, and the feature vectors for the synthetic graph can be sampled accordingly.   
For \texttt{syn-products}, we further restrict the feature sampling to ensure that nodes in the training, validation and test splits are only mapped to nodes in the corresponding splits in the benchmark. This process respects the data splits used in \texttt{ogbn-products}, which are more realistic and challenging than random splits~\cite{hu2020ogb}. For simplicity, in our synthetic benchmarks, all the classes (5 for \texttt{syn-cora} and 10 for \texttt{syn-products} -- Table~\ref{tab:A-synthetic-stats}) are of the same size.

\begin{table}[h]
	\centering
	\caption{Statistics for Synthetic Datasets} %
	\label{tab:A-synthetic-stats}
	{\footnotesize
	\begin{tabular}{lcc}
		\toprule 
		\textbf{Benchmark Name} & \texttt{syn-cora} & \texttt{syn-products} \\
		\midrule
		\textbf{\# Nodes} & 1490 & 10000 \\
		\textbf{\# Edges} & 2965 to 2968  & 59640 to 59648 \\
		\textbf{\# Classes} & 5 & 10 \\
		\textbf{Features} 
		    & \texttt{cora}~\cite{sen2008collective, yang2016revisiting} 
		    & \texttt{ogbn-products}~\cite{hu2020ogb} \\
		\textbf{Homophily $h$} & [0, 0.1, \ldots, 1] & [0, 0.1, \ldots, 1] \\
		\textbf{Degree Range} & 1 to 94 & 1 to 336 \\
		\textbf{Average Degree} & 3.98 & 11.93 \\
		\bottomrule
	\end{tabular}
	}
\end{table}

\paragraph{Experimental setup} For each heterophily ratio $h$ of each benchmark, we independently generate 3 different graphs. For \texttt{syn-cora} and \texttt{syn-products}, we randomly partition 25\% of nodes into training set, 25\% into validation and 50\% into test set. %
All methods share the same training, partition and test splits, and the average and standard derivation of the performance values under the 3 generated graphs are reported as the performance under each heterophily level of each benchmark.

\subsection{Detailed Results on Synthetic Benchmarks}
\label{app:synthetic-results}

Tables~\ref{tab:app-syn-cora-results} and \ref{tab:app-syn-products-results} give the results on \texttt{syn-cora} and \texttt{syn-products} shown in Figure~\ref{fig:5-syn-results} of the main paper (\S~\ref{sec:eval-synthetic}).
Table~\ref{tab:design-ablations} provides the detailed results of the ablation studies that we designed in order to investigate the significance of our design choices, and complements Fig.~\ref{fig:design-ablations} in \S~\ref{sec:eval-synthetic}.

\begin{table}[t]
	\centering
	\caption{\texttt{syn-cora} (Fig.~\ref{fig:syn-cora}): Mean accuracy and standard deviation per method and synthetic dataset (with different homophily ratio $h$). Best method highlighted in gray.} %
	\label{tab:app-syn-cora-results}
    \resizebox{.7\textwidth}{!}{
	\begin{tabular}{lcccccc}
		\toprule 
		\textbf{h} & \textbf{0.00} & \textbf{0.10} & \textbf{0.20} & \textbf{0.30} & \textbf{0.40} & \textbf{0.50}  \\ \midrule %
		\textbf{\method-1} & $77.40{\scriptstyle\pm0.89}$ & $76.82{\scriptstyle\pm1.30}$ & $73.38{\scriptstyle\pm0.95}$ & \cellcolor{gray!15}$75.26{\scriptstyle\pm0.56}$ & $75.66{\scriptstyle\pm2.19}$ & \cellcolor{gray!15}$80.22{\scriptstyle\pm1.35}$ \\ 
		
		\textbf{\method-2} & \cellcolor{gray!15}$77.85{\scriptstyle\pm1.63}$ & \cellcolor{gray!15}$76.87{\scriptstyle\pm0.43}$ & \cellcolor{gray!15}$74.27{\scriptstyle\pm1.30}$ & $74.41{\scriptstyle\pm0.43}$ & \cellcolor{gray!15}$76.33{\scriptstyle\pm1.35}$ & $79.60{\scriptstyle\pm0.48}$ \\
        
		\textbf{GraphSAGE} & $75.97{\scriptstyle \pm1.94}$ & $72.89{\scriptstyle \pm2.42}$ & $70.56{\scriptstyle \pm1.42}$ & $71.81{\scriptstyle \pm0.67}$ & $72.04{\scriptstyle \pm1.68}$ & $76.55{\scriptstyle \pm0.81}$ \\
		\textbf{GCN-Cheby} & $74.23{\scriptstyle \pm0.54}$ & $68.10{\scriptstyle \pm1.75}$ & $64.70{\scriptstyle \pm1.17}$ & $66.71{\scriptstyle \pm1.63}$ & $68.14{\scriptstyle \pm1.56}$ & $73.33{\scriptstyle \pm2.05}$ \\
		\textbf{MixHop} & $62.64{\scriptstyle\pm1.16}$ & $58.93{\scriptstyle\pm2.84}$ & $60.89{\scriptstyle\pm1.20}$ & $65.73{\scriptstyle\pm0.41}$ & $67.87{\scriptstyle\pm4.01}$ & $70.11{\scriptstyle\pm0.34}$ \\
		\midrule
		\textbf{GCN} & $33.65{\scriptstyle \pm1.68}$ & $37.14{\scriptstyle \pm4.60}$ & $42.82{\scriptstyle \pm1.89}$ & $51.10{\scriptstyle \pm0.77}$ & $56.91{\scriptstyle \pm2.56}$ & $66.22{\scriptstyle \pm1.04}$ \\
		\textbf{GAT} & $30.16{\scriptstyle \pm1.32}$ & $33.11{\scriptstyle \pm1.20}$ & $39.11{\scriptstyle \pm0.28}$ & $48.81{\scriptstyle \pm1.57}$ & $55.35{\scriptstyle \pm2.35}$ & $64.52{\scriptstyle \pm0.47}$ \\ 
		\midrule
		\textbf{MLP} & $72.75{\scriptstyle \pm1.51}$ & $74.85{\scriptstyle \pm0.76}$ & $74.05{\scriptstyle \pm0.69}$ & $73.78{\scriptstyle \pm1.14}$ & $73.33{\scriptstyle \pm0.34}$ & $74.81{\scriptstyle \pm1.90}$ \\
		\bottomrule
        \toprule
        \textbf{h} & \textbf{0.60} & \textbf{0.70} & \textbf{0.80} & \textbf{0.90} & \textbf{1.00}\\
        \midrule
        \textbf{\method-1} & $83.62{\scriptstyle\pm0.82}$ & $88.14{\scriptstyle\pm0.31}$ & $91.63{\scriptstyle\pm0.77}$ & $95.53{\scriptstyle\pm0.61}$ & $99.06{\scriptstyle\pm0.27}$ \\
        
        \textbf{\method-2} & \cellcolor{gray!15}$84.43{\scriptstyle\pm1.89}$ & \cellcolor{gray!15}$88.28{\scriptstyle\pm0.66}$ & \cellcolor{gray!15}$92.39{\scriptstyle\pm1.34}$ & \cellcolor{gray!15}$95.97{\scriptstyle\pm0.59}$ & \cellcolor{gray!15}$100.00{\scriptstyle\pm0.00}$ \\

        \textbf{GraphSAGE} & $81.25{\scriptstyle \pm1.04}$ & $85.06{\scriptstyle \pm0.51}$ & $90.78{\scriptstyle \pm1.02}$ & $95.08{\scriptstyle \pm1.16}$  & $99.87{\scriptstyle \pm0.00}$ \\
        \textbf{GCN-Cheby} & $78.88{\scriptstyle \pm0.21}$ & $84.92{\scriptstyle \pm1.03}$ & $90.92{\scriptstyle \pm1.62}$ & $95.97{\scriptstyle \pm1.07}$ & \cellcolor{gray!15} ${100.00{\scriptstyle \pm0.00}}$ \\
        \textbf{MixHop} & $79.78{\scriptstyle\pm1.92}$ & $84.43{\scriptstyle\pm0.94}$ & $91.90{\scriptstyle\pm2.02}$ & $96.82{\scriptstyle\pm0.08}$ & $100.00{\scriptstyle\pm0.00}$ \\
        \midrule
        \textbf{GCN} & $77.32{\scriptstyle \pm1.17}$ & $84.52{\scriptstyle \pm0.54}$ & $91.23{\scriptstyle \pm1.29}$ & $96.11{\scriptstyle \pm0.82}$ & \cellcolor{gray!15} ${100.00{\scriptstyle \pm0.00}}$ \\
  		\textbf{GAT} & $76.29{\scriptstyle \pm1.83}$ & $84.03{\scriptstyle \pm0.97}$ & $90.92{\scriptstyle \pm1.51}$ & $95.88{\scriptstyle \pm0.21}$ & \cellcolor{gray!15} ${100.00{\scriptstyle \pm0.00}}$ \\
  		\midrule
        \textbf{MLP} & $73.42{\scriptstyle \pm1.07}$ & $71.72{\scriptstyle \pm0.62}$ & $72.26{\scriptstyle \pm1.53}$ & $72.53{\scriptstyle \pm2.77}$ & $73.65{\scriptstyle \pm0.41}$ \\
        \bottomrule
	\end{tabular}
}
\end{table}

\begin{table}[t]
	\caption{\texttt{syn-products} (Fig.~\ref{fig:syn-products}): Mean accuracy and standard deviation per method and synthetic dataset (with different homophily ratio $h$). Best method highlighted in gray.} 
	\label{tab:app-syn-products-results}  %
	\centering
    \resizebox{.7\textwidth}{!}{
	\begin{tabular}{lcccccc}
		\toprule 	
		\textbf{h} & \textbf{0.00} & \textbf{0.10} & \textbf{0.20} & \textbf{0.30} & \textbf{0.40} & \textbf{0.50}  \\ \midrule %
		\textbf{\method-1}  & $82.06{\scriptstyle \pm0.24}$ & $78.39{\scriptstyle \pm1.56}$ & $79.37{\scriptstyle \pm0.21}$ & $81.10{\scriptstyle \pm0.22}$ & $84.25{\scriptstyle \pm1.08}$ & $88.15{\scriptstyle \pm0.28}$ \\
		\textbf{\method-2} & $83.37{\scriptstyle \pm0.38}$ & \cellcolor{gray!15} $80.03{\scriptstyle \pm0.84}$ & \cellcolor{gray!15} $81.09{\scriptstyle \pm0.41}$ & \cellcolor{gray!15} $82.79{\scriptstyle \pm0.49}$ & \cellcolor{gray!15} $86.73{\scriptstyle \pm0.66}$ & \cellcolor{gray!15} $90.75{\scriptstyle \pm0.43}$ \\
		\textbf{GraphSAGE} & $77.66{\scriptstyle \pm0.72}$ & $74.04{\scriptstyle \pm1.07}$ & $75.29{\scriptstyle \pm0.82}$ & $76.39{\scriptstyle \pm0.24}$ & $80.49{\scriptstyle \pm0.96}$ & $84.51{\scriptstyle \pm0.51}$\\
		\textbf{GCN-Cheby} & \cellcolor{gray!15} $84.35{\scriptstyle \pm0.62}$ & $76.95{\scriptstyle \pm0.30}$ & $77.07{\scriptstyle \pm0.49}$ & $78.43{\scriptstyle \pm0.73}$ & $85.09{\scriptstyle \pm0.29}$ & $89.66{\scriptstyle \pm0.53}$ \\
		\textbf{MixHop} & $15.39{\scriptstyle\pm1.38}$ & $11.91{\scriptstyle\pm1.17}$ & $14.03{\scriptstyle\pm1.70}$ & $14.92{\scriptstyle\pm0.56}$ & $17.04{\scriptstyle\pm0.40}$ & $18.90{\scriptstyle\pm1.49}$ \\
		\midrule
		\textbf{GCN} & $56.44{\scriptstyle \pm0.59}$ & $51.51{\scriptstyle \pm0.56}$ & $54.97{\scriptstyle \pm0.66}$ & $64.90{\scriptstyle \pm0.90}$ & $76.25{\scriptstyle \pm0.04}$ & $86.43{\scriptstyle \pm0.58}$ \\
		\textbf{GAT} & $27.39{\scriptstyle\pm2.47}$ & $21.49{\scriptstyle\pm2.25}$ & $37.27{\scriptstyle\pm3.99}$ & $44.46{\scriptstyle\pm0.68}$ & $51.86{\scriptstyle\pm8.52}$ & $69.42{\scriptstyle\pm5.30}$ \\
		\midrule
		\textbf{MLP} & $68.63{\scriptstyle \pm0.58}$ & $68.20{\scriptstyle \pm1.20}$ & $68.85{\scriptstyle \pm0.73}$ & $68.65{\scriptstyle \pm0.18}$ & $68.37{\scriptstyle \pm0.85}$ & $68.70{\scriptstyle \pm0.61}$ \\
		\bottomrule
        \toprule
        \textbf{h} & \textbf{0.60} & \textbf{0.70} & \textbf{0.80} & \textbf{0.90} & \textbf{1.00}\\
        \midrule
        \textbf{\method-1} & $92.39{\scriptstyle \pm0.06}$ & $95.69{\scriptstyle \pm0.19}$ & $98.09{\scriptstyle \pm0.23}$ & $99.63{\scriptstyle \pm0.13}$ & $99.93{\scriptstyle \pm0.01}$ \\
        \textbf{\method-2} & $94.81{\scriptstyle \pm0.27}$ & $97.67{\scriptstyle \pm0.18}$ & $99.13{\scriptstyle \pm0.05}$ & $99.89{\scriptstyle \pm0.08}$ & $99.99{\scriptstyle \pm0.01}$ \\
        \textbf{GraphSAGE} & $89.51{\scriptstyle \pm0.29}$ & $93.61{\scriptstyle \pm0.52}$ & $96.66{\scriptstyle \pm0.19}$ & $98.78{\scriptstyle \pm0.11}$ & $99.63{\scriptstyle \pm0.08}$ \\
        \textbf{GCN-Cheby} & \cellcolor{gray!15} $94.99{\scriptstyle \pm0.34}$ & \cellcolor{gray!15} $98.26{\scriptstyle \pm0.11}$ & \cellcolor{gray!15} $99.58{\scriptstyle \pm0.11}$ & \cellcolor{gray!15} $99.93{\scriptstyle \pm0.06}$ & \cellcolor{gray!15} $100.00{\scriptstyle \pm0.00}$ \\
        \textbf{MixHop} & $19.47{\scriptstyle\pm5.21}$ & $21.15{\scriptstyle\pm2.28}$ & $24.16{\scriptstyle\pm3.19}$ & $23.21{\scriptstyle\pm5.30}$ & $25.09{\scriptstyle\pm5.08}$ \\
        \midrule
        \textbf{GCN} & $93.35{\scriptstyle \pm0.28}$ & $97.61{\scriptstyle \pm0.24}$ & $99.33{\scriptstyle \pm0.08}$ & \cellcolor{gray!15}$99.93{\scriptstyle \pm0.01}$  & $99.99{\scriptstyle \pm0.01}$ \\
		\textbf{GAT} & $85.36{\scriptstyle\pm3.67}$ & $93.52{\scriptstyle\pm1.93}$ & $98.84{\scriptstyle\pm0.12}$ & $99.87{\scriptstyle\pm0.06}$ & $99.98{\scriptstyle\pm0.02}$ \\ 
		\midrule
        \textbf{MLP} & $68.21{\scriptstyle \pm0.93}$ & $68.72{\scriptstyle \pm1.11}$ & $68.10{\scriptstyle \pm0.54}$ & $68.36{\scriptstyle \pm1.42}$ & $69.08{\scriptstyle \pm1.03}$ \\
        \bottomrule
	\end{tabular}
}
\end{table}

\begin{table}[t]
	\caption{Ablation studies of \method to show the significance of designs D1-D3 (Fig.~\ref{fig:design-ablations}(a)-(c)): Mean accuracy and standard deviation per method on the \texttt{syn-products} networks.} %
	\label{tab:design-ablations}
	\centering
    \resizebox{.91\textwidth}{!}{
	\begin{tabular}{l@{\hskip 0.8cm}lcccccc}
		\toprule 
		\textbf{Design} & \textbf{h} & \textbf{0.00} & \textbf{0.10} & \textbf{0.20} & \textbf{0.30} & \textbf{0.40} & \textbf{0.50}  \\ 		
		\midrule %
		D1-D3 & \textbf{[S0 / K2] \method-1}  & $82.06{\scriptstyle \pm0.24}$ & $78.39{\scriptstyle \pm1.56}$ & $79.37{\scriptstyle \pm0.21}$ & $81.10{\scriptstyle \pm0.22}$ & $84.25{\scriptstyle \pm1.08}$ & $88.15{\scriptstyle \pm0.28}$ \\
		D3 & \textbf{\method-2} & $83.37{\scriptstyle \pm0.38}$ & $80.03{\scriptstyle \pm0.84}$ & $81.09{\scriptstyle \pm0.41}$ & $82.79{\scriptstyle \pm0.49}$ & $86.73{\scriptstyle \pm0.66}$ & $90.75{\scriptstyle \pm0.43}$ \\
		\midrule
		D1 & \textbf{[NS0] $\mathbf{N_1 + N_2}$} & $52.72{\scriptstyle \pm0.13}$ & $41.65{\scriptstyle \pm0.18}$ & $46.11{\scriptstyle \pm0.86}$ & $58.16{\scriptstyle \pm0.79}$ & $71.10{\scriptstyle \pm0.54}$ & $82.19{\scriptstyle \pm0.40}$ \\
		D1 & \textbf{[NS1] Only $\mathbf{N_1}$} & $40.35{\scriptstyle \pm0.58}$ & $35.17{\scriptstyle \pm0.92}$ & $40.35{\scriptstyle \pm0.92}$ & $52.45{\scriptstyle \pm0.85}$ & $65.62{\scriptstyle \pm0.56}$ & $76.05{\scriptstyle \pm0.38}$ \\
		D1, D2 & \textbf{[S1 / N2] w/o $\neighNoSelfLoop_2$} & $79.65{\scriptstyle \pm0.27}$ & $76.08{\scriptstyle \pm0.76}$ & $76.46{\scriptstyle \pm0.21}$ & $77.29{\scriptstyle \pm0.46}$ & $79.81{\scriptstyle \pm0.88}$ & $83.56{\scriptstyle \pm0.22}$ \\
		D2 & \textbf{[N1] w/o $\neighNoSelfLoop_1$} & $72.27{\scriptstyle \pm0.55}$ & $73.05{\scriptstyle \pm1.23}$ & $75.81{\scriptstyle \pm0.67}$ & $76.83{\scriptstyle \pm0.72}$ & $80.49{\scriptstyle \pm0.72}$ & $82.91{\scriptstyle \pm0.44}$ \\		
	    D2 & \textbf{[N0] w/o 0-hop neighb. (ego)} & $63.55{\scriptstyle \pm0.46}$ & $46.73{\scriptstyle \pm0.42}$ & $42.29{\scriptstyle \pm0.55}$ & $48.20{\scriptstyle \pm0.59}$ & $61.22{\scriptstyle \pm0.35}$ & $75.15{\scriptstyle \pm0.27}$ \\
	    D3 &  \textbf{[K0] No Round-0} & $75.63{\scriptstyle \pm0.19}$ & $61.99{\scriptstyle \pm0.57}$ & $56.36{\scriptstyle \pm0.56}$ & $61.27{\scriptstyle \pm0.71}$ & $73.33{\scriptstyle \pm0.88}$ & $84.51{\scriptstyle \pm0.50}$ \\
	    D3 & \textbf{[K1] No Round-1} & $75.75{\scriptstyle \pm0.90}$ & $75.65{\scriptstyle \pm0.73}$ & $79.25{\scriptstyle \pm0.18}$ & $81.19{\scriptstyle \pm0.33}$ & $84.64{\scriptstyle \pm0.35}$ & $88.46{\scriptstyle \pm0.60}$ \\
	    D3 & \textbf{[R2] Only Round-2} & $73.11{\scriptstyle \pm1.01}$ & $62.47{\scriptstyle \pm1.35}$ & $59.99{\scriptstyle \pm0.43}$ & $64.37{\scriptstyle \pm1.14}$ & $75.43{\scriptstyle \pm0.70}$ & $86.02{\scriptstyle \pm0.79}$ \\
	    \midrule
	    \S~\ref{app:qualitative-comp} & \textbf{Non-linear \method-2 (\S~\ref{app:qualitative-comp})} & $82.23{\scriptstyle \pm0.25}$ & $78.78{\scriptstyle \pm1.04}$ & $80.47{\scriptstyle \pm0.15}$ & $82.08{\scriptstyle \pm0.10}$ & $85.89{\scriptstyle \pm0.53}$ & $89.78{\scriptstyle \pm0.11}$ \\
		\bottomrule \\
        \toprule
        \textbf{Design} & \textbf{h} & \textbf{0.60} & \textbf{0.70} & \textbf{0.80} & \textbf{0.90} & \textbf{0.99} & \textbf{1.00}\\
        \midrule
        D1, D3 & \textbf{[S0 / K2] \method-1} & $92.39{\scriptstyle \pm0.06}$ & $95.69{\scriptstyle \pm0.19}$ & $98.09{\scriptstyle \pm0.23}$ & $99.63{\scriptstyle \pm0.13}$ & $99.88{\scriptstyle \pm0.06}$ & $99.93{\scriptstyle \pm0.01}$ \\
        D3 & \textbf{\method-2} & $94.81{\scriptstyle \pm0.27}$ & $97.67{\scriptstyle \pm0.18}$ & $99.13{\scriptstyle \pm0.05}$ & $99.89{\scriptstyle \pm0.08}$ & $99.98{\scriptstyle \pm0.00}$ & $99.99{\scriptstyle \pm0.01}$ \\
        \midrule
        D1 & \textbf{[NS0] $\mathbf{N_1 + N_2}$} & $90.39{\scriptstyle \pm0.54}$ & $95.25{\scriptstyle \pm0.06}$ & $98.27{\scriptstyle \pm0.13}$ & $99.69{\scriptstyle \pm0.03}$ & $99.98{\scriptstyle \pm0.02}$ & $100.00{\scriptstyle \pm0.00}$ \\
        D1 & \textbf{[NS1] Only $\mathbf{N_1}$} & $84.41{\scriptstyle \pm0.44}$ & $90.15{\scriptstyle \pm0.27}$ & $95.21{\scriptstyle \pm0.34}$ & $97.71{\scriptstyle \pm0.06}$ & $99.56{\scriptstyle \pm0.11}$ & $99.49{\scriptstyle \pm0.11}$ \\
        D1, D2 & \textbf{[S1 / N2] w/o $\neighNoSelfLoop_2$} & $87.39{\scriptstyle \pm0.33}$ & $91.08{\scriptstyle \pm0.50}$ & $94.36{\scriptstyle \pm0.32}$ & $97.01{\scriptstyle \pm0.40}$ & $98.79{\scriptstyle \pm0.23}$ & $98.71{\scriptstyle \pm0.15}$ \\
	    D2 & \textbf{[N1] w/o $\neighNoSelfLoop_1$} & $87.24{\scriptstyle \pm0.21}$ & $92.55{\scriptstyle \pm0.50}$ & $95.64{\scriptstyle \pm0.19}$ & $98.71{\scriptstyle \pm0.13}$ & $99.73{\scriptstyle \pm0.12}$ & $99.83{\scriptstyle \pm0.06}$ \\
        D2 & \textbf{[N0] w/o 0-hop neighb. (ego)} & $86.08{\scriptstyle \pm0.58}$ & $93.03{\scriptstyle \pm0.29}$ & $97.45{\scriptstyle \pm0.09}$ & $99.45{\scriptstyle \pm0.06}$ & $99.98{\scriptstyle \pm0.02}$ & $99.98{\scriptstyle \pm0.03}$ \\
        D3 & \textbf{[K0] No Round-0} & $92.42{\scriptstyle \pm0.13}$ & $96.81{\scriptstyle \pm0.11}$ & $99.09{\scriptstyle \pm0.27}$ & $99.89{\scriptstyle \pm0.01}$ & $100.00{\scriptstyle \pm0.00}$ & $100.00{\scriptstyle \pm0.00}$ \\
        D3 & \textbf{[K1] No Round-1} & $93.05{\scriptstyle \pm0.23}$ & $97.17{\scriptstyle \pm0.36}$ & $99.06{\scriptstyle \pm0.09}$ & $99.89{\scriptstyle \pm0.08}$ & $99.97{\scriptstyle \pm0.02}$ & $99.97{\scriptstyle \pm0.01}$ \\
        D3 & \textbf{[R2] Only Round-2} & $93.79{\scriptstyle \pm0.28}$ & $97.88{\scriptstyle \pm0.18}$ & $99.38{\scriptstyle \pm0.12}$ & $99.89{\scriptstyle \pm0.05}$ & $100.00{\scriptstyle \pm0.00}$ & $100.00{\scriptstyle \pm0.00}$ \\
        \midrule
        \S~\ref{app:qualitative-comp} & \textbf{Non-linear \method-2 } & $93.68{\scriptstyle \pm0.50}$ & $96.73{\scriptstyle \pm0.23}$ & $98.55{\scriptstyle \pm0.06}$ & $99.74{\scriptstyle \pm0.05}$ & $99.96{\scriptstyle \pm0.04}$ & $99.93{\scriptstyle \pm0.03}$ \\
        \bottomrule
	\end{tabular}
}
\end{table}

\clearpage
\section{Real Datasets: Details}
\label{app:real}

\paragraph{Datasets}  In our experiments, we use the following real-world datasets with varying levels of homophily ratios $h$. Some network statistics are given in Table~\ref{tab:5-real-results}.
\begin{itemize}
\item \textbf{Texas, Wisconsin and Cornell} are graphs representing links between web pages of the corresponding universities, originally collected by the CMU WebKB project. We used the preprocessed version  in \cite{Pei2020Geom-GCN}. In these networks, nodes are web pages, which are classified into 5 categories: course, faculty, student, project, staff. 
\item \textbf{Squirrel and Chameleon} are subgraphs of web pages in Wikipedia discussing the corresponding topics, collected by \cite{rozemberczki2019multiscale}. For the classification task, we utilize the class labels generated by \cite{Pei2020Geom-GCN}, where the nodes are categorized into 5 classes based on the amount of their average traffic.  
\item \textbf{Actor} is a graph representing actor co-occurrence in Wikipedia pages, processed by \cite{Pei2020Geom-GCN} based on the film-director-actor-writer network in \cite{tang2009social-fc}. We also use the class labels generated by \cite{Pei2020Geom-GCN}. %
\item \textbf{Cora, Pubmed and Citeseer} are citation graphs originally introduced in \cite{sen2008collective, namata2012query}, which are among the most widely used benchmarks for semi-supervised node classification \cite{shchur2018pitfalls, hu2020ogb}. Each node is assigned a class label based on the research field.
These datasets use a bag of words representation as the feature vector for each node.
\item \textbf{Cora Full} is an extended version of Cora, introduced in \cite{bojchevski2018deep, shchur2018pitfalls}, which contain more papers and research fields than Cora. This dataset also uses a bag of words representation as the feature vector for each node. 
\end{itemize}

\paragraph{Data Limitations} As discussed in \cite{shchur2018pitfalls, hu2020ogb}, Cora, Pubmed and Citeseer are widely adopted as benchmarks for semi-supervised node classification tasks; however, all these benchmark graphs display strong homophily, with edge homophily ratio $h \geq 0.7$. As a result, the wide adaptation of these benchmarks have masked the limitations of the homophily assumption in many existing GNN models. 
Open Graph Benchmark is a recent effort of proposing more challenging, realistic benchmarks with improved data quality comparing to the existing benchmarks \cite{hu2020ogb}. However, with respect to homophily, we found that the proposed OGB datasets display homophily $h > 0.5$. 

In our synthetic experiments (\S~\ref{app:synthetic}), we used \texttt{ogbn-products} from this effort to generate higher quality synthetic benchmarks while varying the homophily ratio $h$. 
In our experiments on real datasets, we go beyond the typically-used benchmarks (Cora, Pubmed, Citeseer) and consider benchmarks with strong heterophily (Table~\ref{tab:5-real-results}). That said, these datasets also have limitations, including relatively small sizes (e.g., WebKB benchmarks), artificial classes (e.g., Squirrel and Chameleon have class labels based on ranking of page traffic), or unusual network structure (e.g., \texttt{Squirrel} and \texttt{Chameleon} are dense, with many nodes sharing the same neighbors --- cf. \S~\ref{sec:real-eval}). 
We hope that this paper will encourage future work on more diverse datasets with different levels of homophily, and lead to higher quality datasets for benchmarking GNN models in the heterophily settings.

\end{document}